\documentclass[letterpaper,twocolumn,10pt]{article}
\usepackage{usenix2019_v3}
\usepackage{tikz}
\usepackage{amsmath}
\usepackage{multirow}
\usepackage{makecell}

\usepackage{filecontents}
\usepackage{subcaption}
\usepackage{xspace}
\usepackage{booktabs}
\usepackage{xurl}

\usepackage{subcaption}
\usepackage{hyperref}
\usepackage{cleveref}
\usepackage{enumitem}
\usepackage{fdsymbol}
\usepackage{flushend}

\begin{document}

\date{}

\newcommand{\systemname}[0]{\textsc{PixelMod}\xspace}
\newcommand{\done}[0]{$\textbf{D}_1$\xspace}
\newcommand{\dtwo}[0]{$\textbf{D}_2$\xspace}
\newcommand{\dthree}[0]{$\textbf{D}_3$\xspace}
\newcommand{\visualgroundtruth}[0]{$\textbf{GT}_{viz}$\xspace}

\title{\systemname: Improving Soft Moderation of\\ Visual Misleading Information on Twitter\thanks{This paper is accepted for publication at the 2024 USENIX Security Symposium. Please cite accordingly.}}

\author{Pujan Paudel$^\diamondsuit$, Chen Ling$^\diamondsuit$, Jeremy Blackburn$^\clubsuit$, and Gianluca Stringhini$^\diamondsuit$\\
$^\diamondsuit$Boston University, $^\clubsuit$Binghamton University\\
\{ppaudel,ccling,gian\}@bu.edu, jblackbu@binghamton.edu
}

\maketitle

\newcommand{\descr}[1]{\smallskip\noindent\textbf{#1}}
\newcommand{\descrit}[1]{\smallskip\noindent\emph{#1}}

\newif
\ifcomment
\commentfalse
\ifcomment
\newcommand{\gs}[1]{{\bf \textcolor{red}{GS: #1}}}
\newcommand{\ppaudel}[1]{{\bf \textcolor{blue}{PP: #1}}}
\newcommand{\jbnote}[1]{{\bf \textcolor{yellow}{JB: #1}}}
\newcommand{\cici}[1]{{\bf \textcolor {green}{CL: #1}}}
\newcommand{\revision}[1]{{\textcolor {magenta}{#1}}}
\else
\newcommand{\gs}[1]{}
\newcommand{\ppaudel}[1]{}
\newcommand{\jbnote}[1]{}
\newcommand{\cici}[1]{}
\newcommand{\revision}[1]{{\textcolor {magenta}{#1}}}

\fi

\begin{abstract}
Images are a powerful and immediate vehicle to carry misleading or outright false messages, yet identifying image-based misinformation at scale poses unique challenges.
In this paper, we present \systemname, a system that leverages perceptual hashes, vector databases, and optical character recognition (OCR) to efficiently identify images that are candidates to receive soft moderation labels on Twitter.
We show that \systemname outperforms existing image similarity approaches when applied to soft moderation, with negligible performance overhead.
We then test \systemname on a dataset of tweets surrounding the 2020 US Presidential Election, and find that it is able to identify visually misleading images that are candidates for soft moderation with 0.99\% false detection and 2.06\% false negatives.

\end{abstract}

\maketitle

\section{Introduction}
Online users are constantly bombarded with false and misleading information, whether it is inaccurate and shared without malice (i.e., \emph{misinformation}), or deliberately crafted to achieve a malicious goal, for example by state actors (i.e., \emph{disinformation})~\cite{saeed2022trollmagnifier,starbird2019disinformation}. 
To help their users better distinguish between accurate and misleading or false information, online platforms like Twitter have introduced \emph{soft moderation} measures, where they add labels to posts that include inaccurate information, with the goal of providing better context to these claims.
Unfortunately, little is known on how these moderation decisions are made by the platforms, and recent work found glaring issues with Twitter's soft moderation approach~\cite{lawfare_2022,zannettou2021won}.

\begin{figure}[t]{}{}
     \centering
     \begin{subfigure}[b]{0.3\columnwidth}
         \centering
         \includegraphics[width=\textwidth]{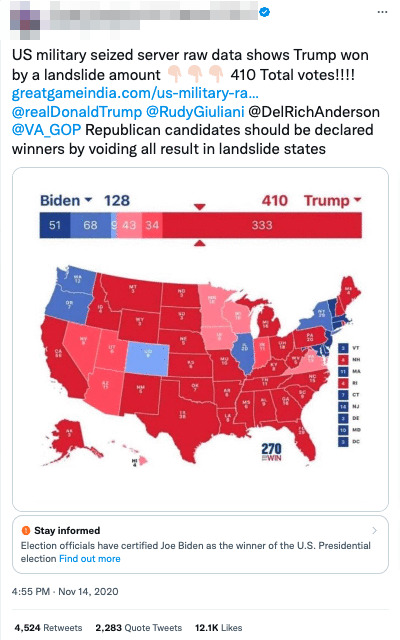}
         \caption{Example of a moderated tweet containing an image}
         \label{fig:tweet_moderated_img}
     \end{subfigure}
     \hfill
     \begin{subfigure}[b]{0.3\columnwidth}
         \centering
         \includegraphics[width=\textwidth]{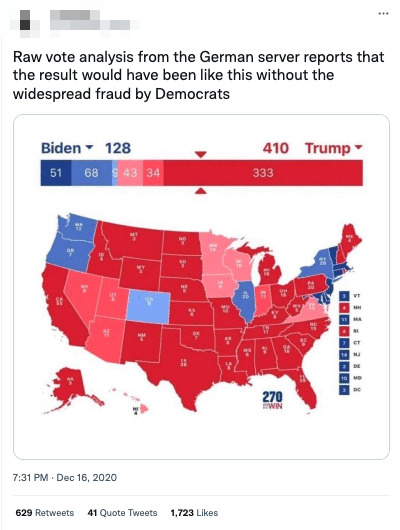}
         \caption{Example of an unmoderated tweet containing the same image}
         \label{fig:tweet_unmoderated1_img}
     \end{subfigure}
     \hfill
     \begin{subfigure}[b]{0.3\columnwidth}
         \centering
         \includegraphics[width=\textwidth]{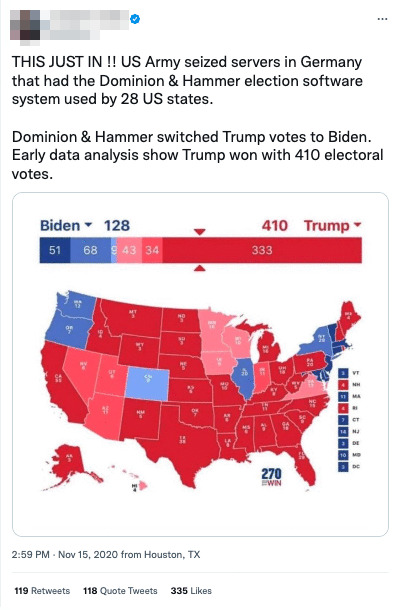}
         \caption{Example of another unmoderated tweet containing the same image}
         \label{fig:tweet_unmoderated2_img}
     \end{subfigure}
    \caption{Three tweets discussing a false electoral vote map showing Trump landslide victory based on false reports of seized election servers in Germany. Twitter added a warning label only to the first tweet.}
        \label{fig:tweet_moderation_examples_img}
\end{figure}

While previous work on automated soft moderation~\cite{paudel2022lambretta} builds a solid foundation for the soft moderation of textual content, false information is not only spread through text, but images are commonly used to convey false and misleading narratives~\cite{garimella2017image,matatov2022stop,wang2023understanding,zannettou2019characterizing}.
While Twitter applies soft moderation decisions to posts containing images, the way in which this is done is not publicly known, and a preliminary analysis that we conducted shows that these labels are not applied uniformly and pervasively.
For example, in Figure~\ref{fig:tweet_moderation_examples_img}, we show three tweets that contain identical images discussing the same debunked electoral map claiming Trump's landslide victory with 410 votes after elections servers were seized in Germany by the U.S military~\cite{usatoday_410}.
The first tweet (Figure~\ref{fig:tweet_moderated_img}) received a warning label, while the other two tweets containing the same image (Figures~\ref{fig:tweet_unmoderated1_img} and~\ref{fig:tweet_unmoderated2_img}) did not receive any intervention, despite discussing the same misleading narrative. 

These observations highlight the need for effective automated approaches to identify image content that should receive soft moderation.
Moderating images, however, poses unique challenges.
First, the sheer amount of data published on social media makes this a particularly challenging task, especially on platforms like Twitter, where 50 million tweets are posted every day~\cite{twitternumbers}.
Previous work has dealt with this problem by adapting \emph{perceptual hashing} algorithms~\cite{garimella2017image,wang2020understanding,zannettou2019characterizing}.
These algorithms identify visually similar images by leveraging syntactic embeddings, which are lightweight to calculate.
Even if the calculation of hashes is relatively cheap, comparing an image against a large number of potential candidates is inefficient and does not scale.
In addition to the computational costs of comparing images, previous research has shown that image-based false information is highly contextual, and that similar images can be misinformation, satire, or even completely benign based on the context in which they are used~\cite{wang2023understanding,wang2020understanding}.
Existing approaches using perceptual hashing lack the nuance to determine this context, making them prone to false positives and unfit for the misinformation moderation use case.

\descr{Technical Roadmap.}
In this paper, we aim to address the limitations of previous approaches and develop a scalable, performant, and effective system able to analyze images posted on social media and identify candidates for soft moderation. 
We present \systemname, an image search system designed to assist platform moderators by flagging visual content that is similar to content they have previously applied moderation labels to. 
\systemname takes as seed input a list of images initially moderated by Twitter and encodes the images using perceptual hashing~\cite{davis2019open,monga2006perceptual}.
To allow efficient hash comparisons, \systemname leverages Milvus~\cite{wang2021milvus}, a vector database that is optimized for searching among millions of hash records.
After finding images that are visually similar, \systemname leverages OCR to identify the context in which an image is used, allowing it to rule out false positives.

\systemname successfully identifies both tweets in Figures~\ref{fig:tweet_unmoderated1_img} and~\ref{fig:tweet_unmoderated2_img} when queried with the image in Figure~\ref{fig:tweet_moderated_img}.
Additionally, \systemname not only identifies images identical to the seed images but also visual variants of the images such as memes and fauxtography, which are commonly used for spreading misleading information~\cite{garimella2020images,wang2020understanding,zannettou2018origins}.

\descr{Main Contributions \& Findings.}
\systemname addresses the limitations of previous image-based moderation approaches proposed in the academic literature, 
is platform-independent, and only requires a single misleading image already intervened by human moderators.
With the ability to identify thousands of posts spreading misleading images, we show that an image search system like \systemname can be used as a foundation for large-scale soft moderation intervention system for images.
While a large amount of research has been conducted on perceptual hashing and OCR, the novelty of our work lies in combining and tuning the two into a working end-to-end soft moderation system which achieves a much larger coverage than perceptual hashing alone.
Besides content moderation, the takeaways offered by our work can be helpful for researchers and practitioners in building image similarity search systems for related security problems like phishing~\cite{kintis2017hiding} and CSAM detection~\cite{zoom_csam,google_csam}.

We compare \systemname with alternative approaches to assess visual similarity, showing that \systemname outperforms them in terms of F1-score.
In particular, the use of OCR to determine the context of false images allows \systemname to operate using a higher similarity threshold than existing approaches only based on perceptual hashing, increasing recall while keeping precision high as well, suffering only a negligible runtime performance hit in return.

We then test \systemname on Twitter data discussing the 2020 US Presidential Election.
Starting from a seed set of 959 tweets containing misleading images, we use \systemname to retrieve 40,244 tweets in the wild that are candidates for moderation.
Performing a thorough manual analysis, we find \systemname has a false negative rate of 2.06\% and a false positive rate of 0.99\%.
These findings are a positive step towards building automated systems that complement Twitter's existing systems to moderate misleading images and improve the state of content moderation. 

To better understand the way in which soft moderation was adopted by Twitter during that period, we perform a qualitative measurement study of the content flagged by \systemname, comparing it to tweets that received soft moderation labels from Twitter.
We categorize the images identified by \systemname based on Twitter's Platform Policy, finding that although different types of platform policies were moderated at different rates by Twitter, none of the policy violations exceeded a moderation rate of 5.96\%.

\begin{figure*}{}{}
  \centering
    \includegraphics[width=\linewidth]{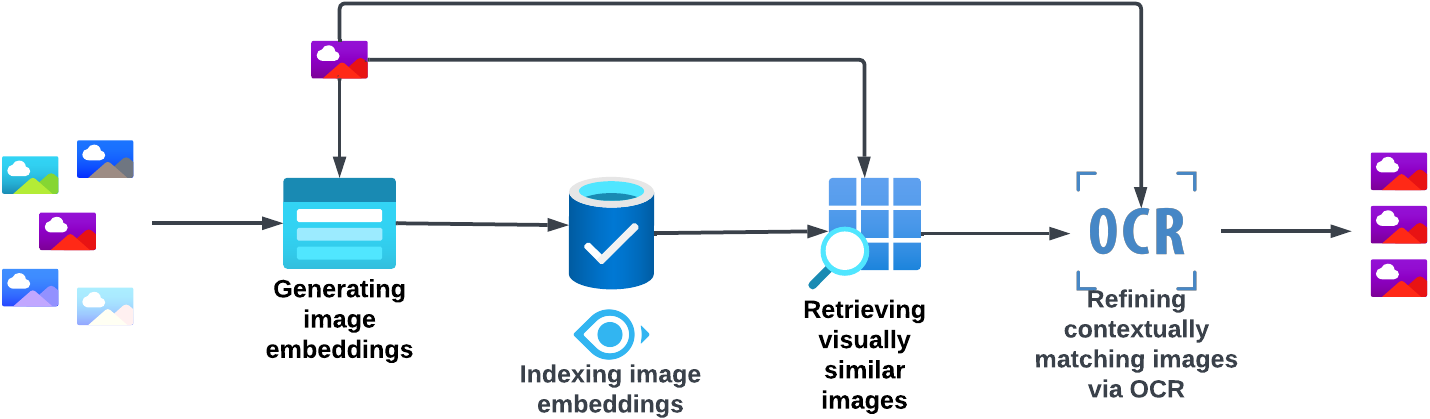}
	\caption{Overview of our image analysis pipeline.} 
  \label{fig:image_processing_pipeline}
\end{figure*}

\section{Overview of \systemname}
\label{sec:overview}
\systemname takes a \emph{query image} that a platform wants to investigate for further moderation actions and retrieves images that are both visually and contextually similar.
This allows a platform to quickly identify a large, high-certainty candidate list of posts for moderation.
\systemname has four stages, as illustrated by Figure~\ref{fig:image_processing_pipeline}: i)~\emph{generating} image embeddings, ii)~\emph{indexing} image embeddings, iii)~\emph{retrieving} visually similar images through approximate nearest neighbor search, and iv)~\emph{refining} contextually matching images via Optical Character Recognition (OCR).
We demonstrate \systemname in the context of Twitter, identifying visually misleading information related to the 2020 US Presidential Election.
However, \systemname's architecture is designed to be portable to any social media platform.

\subsection{Background}
Identifying visually similar images given a query image is a reverse image search problem, which falls under the broader research area of Content-Based Image Retrieval systems~\cite{gudivada1995content}.
The goal is to efficiently collect, index, and search for visually similar images over an index
of millions of images, and there are several publicly available solutions (e.g., TinEye~\cite{tineye}, Google Vision AI~\cite{google_cloud_vision}).
We conducted a preliminary exploration of these systems and identified two major limitations: 1)~they require a paid subscription for programmatic access, and 2)~they perform poorly for returning images from Twitter, Facebook, and other social media platforms because of platform restrictions on their crawlers.
To address these limitations, we built a custom image search system focused exclusively on social media that collects images, generates syntactic embeddings (pHash and PDQHash), and indexes them (using the Milvus~\cite{wang2021milvus} vector database).

\descr{Motivating the need for multi-modality in visual similarity search.}
Prior works have discussed that incorporating contextual information within images is important when identifying when they have been manipulated or miscaptioned~\cite{wang2020understanding}.
Zanettou et al.~\cite{zannettou2018origins} discuss similar existing shortcomings of image similarity systems and the need of incorporating semantic components like OCR in image detection pipelines to effectively identify image memes.
Another work studying misinformation images on WhatsApp in India~\cite{garimella2020images} used semantic features like OCR from Google Cloud Vision to identify differences between different types of misinformation images (images taken out of context, photoshopped images, and memes) after retrieving images visually similar to fact-checked images.
However, no existing work uses contextual information as a part of the detection system or pipeline itself.

\begin{figure}[t]{}{}
     \centering
     \begin{subfigure}[b]{0.31\linewidth}
         \centering
    \includegraphics[width=\linewidth]{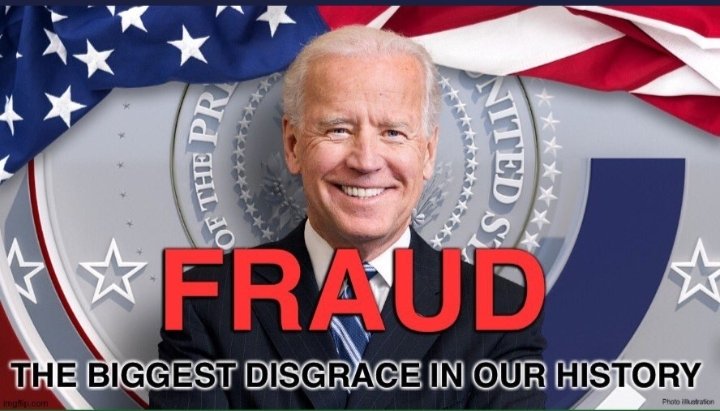}
  \caption{\emph{Query} image}
  \label{fig:context_motivator_1}
     \end{subfigure}
     \hfill
     \begin{subfigure}[b]{0.31\linewidth}
           \centering
    \includegraphics[width=\linewidth]{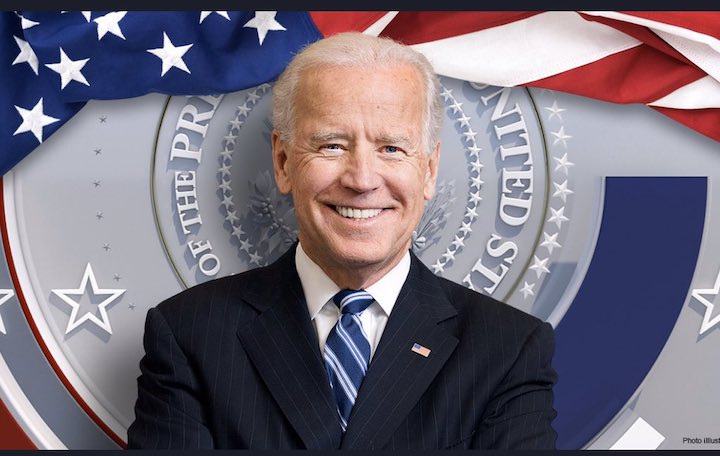}
  \caption{Visually similar match \#1}
  \label{fig:context_motivator_2}
     \end{subfigure}
     \hfill
     \begin{subfigure}[b]{0.31\linewidth}
         \centering
    \includegraphics[width=\linewidth]{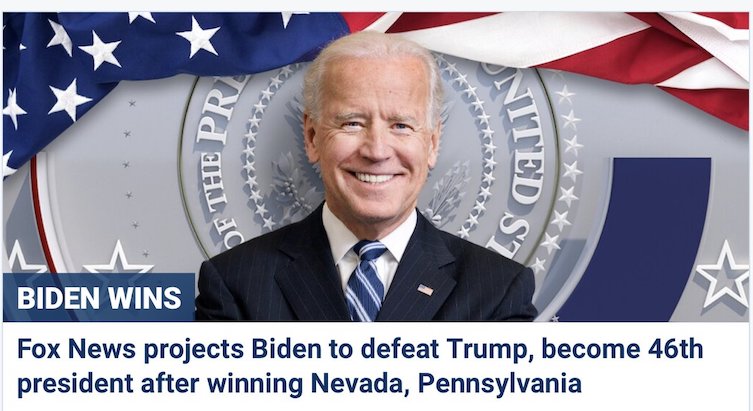}
  \caption{Visually similar match \#2}
  \label{fig:context_motivator_3}
     \end{subfigure}
    \caption{Example query image and results retrieved.}
        \label{fig:context_motivator}
\end{figure}

Existing detection systems can identify images that are very similar to misleading images upon querying, but are unable to verify if the identified images are used in the same misleading context as the query image.
We illustrate this case with an example in Figure~\ref{fig:context_motivator}.
When we query existing image similarity systems with Figure~\ref{fig:context_motivator_1}, they will typically return Figure~\ref{fig:context_motivator_2}, and Figure~\ref{fig:context_motivator_3} as matching results with strong confidence.
However, Figure~\ref{fig:context_motivator_2}, and Figure~\ref{fig:context_motivator_3} are not related to the query image under question, which contains the text of ``Fraud. The Biggest Disgrace In Our History'', despite being visually similar to the query.
Figure~\ref{fig:context_motivator_2} is an image used in a different context, with no connotation to the narrative of ``Election Fraud'' during the 2020 Elections, whereas Figure~\ref{fig:context_motivator_3} is used in another context: Fox News projecting a win for Joe Biden.
Thus, the results returned are used in entirely different context than the query image, lacking the connotation of ``Election Fraud'' which makes the query image misleading.
A similar problem occurs when dealing with \emph{fauxtography}, where images are manipulated with the inclusion of visual changes, overlay text, or are used out of context with the goal of misleading the user~\cite{wang2020understanding}.
This highlights the need for incorporating context when using image similarity systems to improve misinformation detection systems.

To address this issue, \systemname analyzes the overlay text included in images by using optical character recognition (OCR) to better capture the context in which an image is used and then incorporate this context as a part of our image similarity pipeline.
Including this context also helps overcome existing limitations of prior work using perceptual hashes that have to rely on lower similarity thresholds for matching~\cite{garimella2020images,wang2020understanding}.
Incorporating context allows us to increase image similarity thresholds while at the same time achieving better recall without hurting precision.

\subsection{Generating image embeddings}
There are two types of embeddings we can extract from images for retrieval purposes.
The first, \emph{syntactic embeddings}, are fingerprints of an image's visual features, computed such that two visually similar images will have identical, or very similar fingerprints.
The second, \emph{semantic embeddings}, capture features within the images (for example, if a given image is a chart or a portrait), and match images that are visually dissimilar, but still have similar meanings.

For our purposes, syntactic embeddings are more useful for two reasons.
First, images identified through them are minor variations of the query image, making them easy to interpret and take action on.
Second, semantic embeddings use complex deep learning architectures~\cite{he2016deep,simonyan2014very}, while syntactic embeddings are relatively fast and cheap to train and use.
This means we can scale to millions of images with relatively few resources (e.g., \systemname does not require a GPU.)

We generate our syntactic embeddings via \emph{perceptual hashing}.
We make use of two different \emph{perceptual hashing} embeddings: i)~pHash~\cite{monga2006perceptual}, and ii)~PDQHash~\cite{davis2019open}.
pHash encodes images as 64-bit vectors by extracting geometry-preserving feature points using Discrete Cosine Transform (DCT).
The resulting vectors are invariant to simple image transformations and minor coloring differences.
PDQHash is an improvement over pHash, encoding images as 256-bit vectors.
Prior work studying the spread of misleading images in social media has used both pHash and PDQHash~\cite{javed2020first,kazemi2021tiplines,reis2020can,wang2020understanding,zannettou2019characterizing}.
We evaluate pHash and PDQHash with respect to precision, recall, as well as the time taken to generate an embedding, index it, and eventually query the index.

\subsection{Indexing image embeddings}
The most similar image to a query is the one whose embedding is the shortest distance from the query image's embedding.
While we could perform a pairwise comparison with the query image and all candidate images, this is intractable at the scale at which social media platforms operate.
To address this, there are specialized vector databases that can be used to perform queries over embeddings.
For \systemname, we chose to use Milvus~\cite{wang2021milvus} a purpose-built vector management system.
Milvus abstracts away lower-level details (e.g., index creation and item insertion) via a higher-level API and Facebook's Faiss~\cite{johnson2019billion} library for efficient dense vector similarity search.

Milvus has two types of indexes: i)~\texttt{BIN\_FLAT} and ii)~\texttt{BIN\_IVF\_FLAT}.
\texttt{BIN\_FLAT} guarantees a 100\% recall rate by exhaustively searching for matches to a query embedding.
\texttt{BIN\_IVF\_FLAT} is quantization-based, dividing embeddings into multiple cluster units and comparing the query embedding to the center of each cluster.
\texttt{BIN\_IVF\_FLAT} is faster than \texttt{BIN\_FLAT} but requires tuning hyperparameters to find an optimal recall rate.
Milvus recommends using \texttt{BIN\_FLAT} for datasets in the order of a few million vectors, which fits the datasets we use to evaluate \systemname

\subsection{Retrieving visually similar images}
Vector databases like Milvus scale by leveraging approximate nearest neighbor search across their indices.
Milvus determines the nearest neighbors by scoring candidates with a similarity metric.
We use Hamming distance, leaving us with the next challenge of choosing an appropriate threshold to treat as ``visually similar.''
We select Hamming distance over cosine similarity as prior work on perceptual hashing have predominantly used this metric to assess visual similarity on both PHash and PDQHash ( e.g. baseline systems~\cite{wang2020understanding} and ~\cite{zannettou2018origins}).
Additionally, we index the perceptual hashes on Milvus as binary hashes, and using Hamming distance as the similarity metric allows us to efficiently compare the binary representations of images.
Previous works using pHash to study misinformation on a variety of social media platforms have recommended a Hamming distance of six~\cite{wang2020understanding}, and eight~\cite{zannettou2018origins}.
The developers of PDQhash suggest a distance of 32, which has been used to study misinformation on WhatsApp~\cite{kazemi2021tiplines,reis2020can}.
Other work using PDQHash has used a threshold of 40 to study the spread of COVID-19-related media in Pakistan over WhatsApp~\cite{javed2020first}.
Ultimately, choosing a threshold depends on the dataset used and the particulars of how ``visual similarity'' is defined.
In Section~\ref{sec:system_evaluation}, we experimentally evaluate the choice of the visual similarity threshold ($\theta_{visual}$) by optimizing over both precision and recall.

\subsection{Refining matched images using OCR}
The final component of \systemname involves processing the results retrieved as ``visually similar'' to the query image by comparing text extracted via an Optical Character Recognition (OCR) engine.
For the rest of the paper, we refer to the output text from an image extracted from an OCR engine as OCR label.
The advantages of applying this OCR driven post-processing to the matched results is two-fold: i)~it eliminates the limitations of underlying perceptual hashing algorithms by filtering out their false positives and ii)~it allows us to explore relatively larger distance thresholds for visual similarity matching, thus allowing us to minimize false negatives.
Prior applications of perceptual hashing algorithms for identifying visually similar images have used conservative thresholds~\cite{kazemi2021tiplines,javed2020first,wang2020understanding} to minimize false positives.
In contrast, we aim to exploit larger thresholds to increase our recall as well as have a more tuneable mechanism with respect to precision trade-offs. 
In Section~\ref{sec:system_evaluation}, we first validate the choice of Google Cloud Vision API as the underlying OCR engine~\cite{google_cloud_vision}, and experimentally evaluate the calibration of the similarity metric for the extracted OCR labels as our OCR engine) and corresponding threshold for similarity ($\theta_{textual}$).

\section{Datasets}
\label{sec:datasets}

We use three different datasets to evaluate \systemname (summarized in Table~\ref{tab:statstable}).
Our first dataset, \done, contains 7.6M tweets collected via the Twitter Streaming API using a set of 2020 US Election hashtags (e.g., \#ballotfraud, \#voterfraud, \#electionfraud, \#stopthesteal) curated by Abilov et al.~\cite{abilov2021voterfraud2020}.
To comply with Twitter's terms of service, this dataset only provides tweet IDs, therefore, we need to retrieve the full tweet content from the Twitter API through a process called \emph{hydration}.
Out of the 4,017,259 tweets we were able to rehydrate, about 218K have at least one ``media'' entity: either an image or a video (tweets with videos also contain a thumbnail image, which we include in our dataset).

Our second dataset, \dtwo, consists of 13M tweets with media collected via the 1\% Twitter stream from November 1 to December 31, 2020.
Unlike \done, we do not filter any keywords, therefore \dtwo is a uniform 1\% sample of Twitter.

Finally, we collect \dthree via Twitter's Academic Research full-archive search endpoint in late 2022, by querying \emph{context annotations}~\cite{tweet_annotations} associated with the 2020 US Elections.
Twitter annotates tweets with context annotations by semantically analyzing their content and metadata, categorizing them into a nested structure of domains and entity labels.
The more than 80 domains cover things like politics and TV shows, while the nearly 145K entities cover details of the domains ranging from elections to festivals, to politics and media personalities.

We identify two context annotations related to the 2020 election by retrieving the context annotations of all tweets from \done: 1)~``2020 US Election Day'' and 2)~``2020 US Presidential Election,'' both in the \emph{Events} domain.
We also found other context annotations from the tweets, but they were associated with ``Political figures'' and ``Politics'' in general, not the 2020 elections specifically.
To minimize the noise that can be introduced when using such a broad context, we do not include those annotations when building \dthree.
In the end, \dthree consists of the 6.9M tweets that we retrieve using the ``2020 US Election Day'' and ``2020 US Presidential Election'' context annotations and limiting our search to November 1st, 2020 to December 31st, 2020.

\descr{Index building.}
We build two different Milvus indices for pHash and PDQHash embeddings, using all the images in our datasets.
The index size for both types of embeddings is 19.7M, after combining all the tweets from \done, \dtwo, and \dthree.

\begin{table}[t]
\centering
\small
    \setlength{\tabcolsep}{4pt}
\begin{tabular}{cc}
\toprule
\textbf{Dataset} & \textbf{\# Tweets with Images} \\ \midrule
VoterFraud Dataset (\done~)               &  217,868                     \\
1\% Twitter Stream (\dtwo~)         &  12,560,319              \\
Twitter Context Annotations (\dthree~)   &   6,952,300 \\  \bottomrule
\end{tabular}
\caption{Overview of our dataset.}
\label{tab:statstable}
\end{table}

\section{System Validation}
\label{sec:system_evaluation}

In this section, we present the validation strategy for \systemname.
First, we discuss how we build a ground truth dataset. 
Then, we describe our experimental setting to determine the best hashing mechanism, and the best string similarity algorithm for the OCR labels, and the corresponding $\theta_{visual}$ and $\theta_{textual}$ to use with the hashing algorithms and string similarity algorithms.
The goal is to experimentally determine the best operating values for \systemname by optimizing over both precision and recall for the overall system.

\descr{Building ground truth.}
First, we randomly sample 50 images from \done.
For each of these 50 images, we query both the pHash and PDQHash indexes for similar images, limiting the results to the maximum threshold for each algorithm (10 and 90).
We select 90 as the maximum threshold for PDQHash as its developers experimentally verified it as the upper bound for images that are known to be different.
For pHash, we select 10 as the maximum threshold because previous work has found that higher thresholds produce results that are too noisy~\cite{zannettou2018origins}.
This results in 11,825 unique images retrieved as similar across both indices for the same set of 50 \emph{query images}.
Then we manually annotate the results using a pairwise image annotation tool developed by~\cite{dutta2016vgg}.
Note that the goal of this annotation was to verify that images were \emph{visually} and \emph{contextually} similar to the original one (i.e., their overlay text contained similar words).
For this reason, we did not need to build a codebook and have multiple annotators agree on the results as we did for more subjective experiments later in the paper.
In the end, we find 9,785 images that are similar to our 50 source images.
We refer to these 9,785 images as~\visualgroundtruth.

\descr{Determining $\theta_{visual}$}. 
Next, we use the images in \visualgroundtruth to determine the accuracy of the results returned by the two hashing algorithms when using different thresholds for $\theta_{visual}$.
Our Milvus indexes return the closest set of embeddings for a query image, scored by the Hamming distance to the candidate image.
For pHash, we experiment with a range of Hamming distance thresholds from \textit{4} to \textit{10}, which is in line with previous works that found an optimal threshold between \textit{6}~\cite{wang2020understanding} and \textit{8}~\cite{zannettou2018origins}.
For PDQhash, on the other hand, it is recommended to use a range of pairwise distances for identifying similar images~\cite{davis2019open}.
Therefore, we test multiple threshold ranges, some used by previous work using PDQHash~\cite{javed2020first,kazemi2021tiplines,reis2020can} (\textit{32}, \textit{48}) as well as three additional ranges to take us up to the maximum possible threshold (\textit{64}, \textit{80}, and \textit{90}).

\descr{Validating Google Cloud Vision API for OCR.}
To validate the ability of Google Cloud Vision API to correctly extract the OCR labels in a query image, 
we sample 50 images from \visualgroundtruth and create a ground truth of the text contained in the images.
These are examples of image occurring ``in the wild,'' which contain text in specialized fonts, small text in lower quality images, or text in artistic fonts, and are therefore a good test for the OCR component of \systemname.
On this dataset, the median Jaccard similarity of the ground truth text and the one extracted by Cloud Vision OCR is 1.0, and the mean is 0.95.
This validates that Google Cloud Vision API can be succesfully used as the underlying engine for identifying the contextual text contained in the images of \systemname.
Upon doing some error analysis of the OCR text, we find that the output of the Cloud Vision API is not missing any text present in the images, but is sometimes parsed in different order than the ground truth.
This is an artifact of how the OCR engine works on different regions of the image.
We argue that this would not pose a problem when using the system for \systemname as the method will work consistently across all the source images and the potential matches retrieved through perceptual hashing.

\descr{Determining $\theta_{textual}$}.
After validating that Cloud Vision API can be succesfully used to extract the contextual text contained in images, we aim to determine the appropriate text similarity metric and corresponding similarity threshold to compare the OCR labels of the retrieved visual matches with the \textit{query} images.
We first check if the \textit{query} image contains any OCR label, which we call $label_{query}$, using Google Cloud Vision API.
If $label_{query}$ is not empty, then, for each visually matching image, we compute the corresponding OCR labels ($label_{match}$) using the same API endpoint.
We experiment with multiple text similarity metrics, and multiple similarity thresholds to assess if $label_{match}$ is similar to $label_{query}$.
The similarity metrics we experiment with are: i)~Normalized Levenshtein similarity, ii)~Jaro-Winkler similarity, iii)~similarity metric based on Longest Common Subsequence (Metric LCS), and iv)~Jaccard-index similarity.
For each similarity metric, we measure F1 scores on the similarity threshold range of $0,0.05, 0.1, 0.15, \dots, 0.75, 0.8$
Note that, for Jaccard index, we experiment with values of n-grams ranging from 1 to 5 as this algorithms convert string into  a set of n-grams when computing the similarity. 

\begin{figure}{}{}
 \centering
   \includegraphics[width=\linewidth]{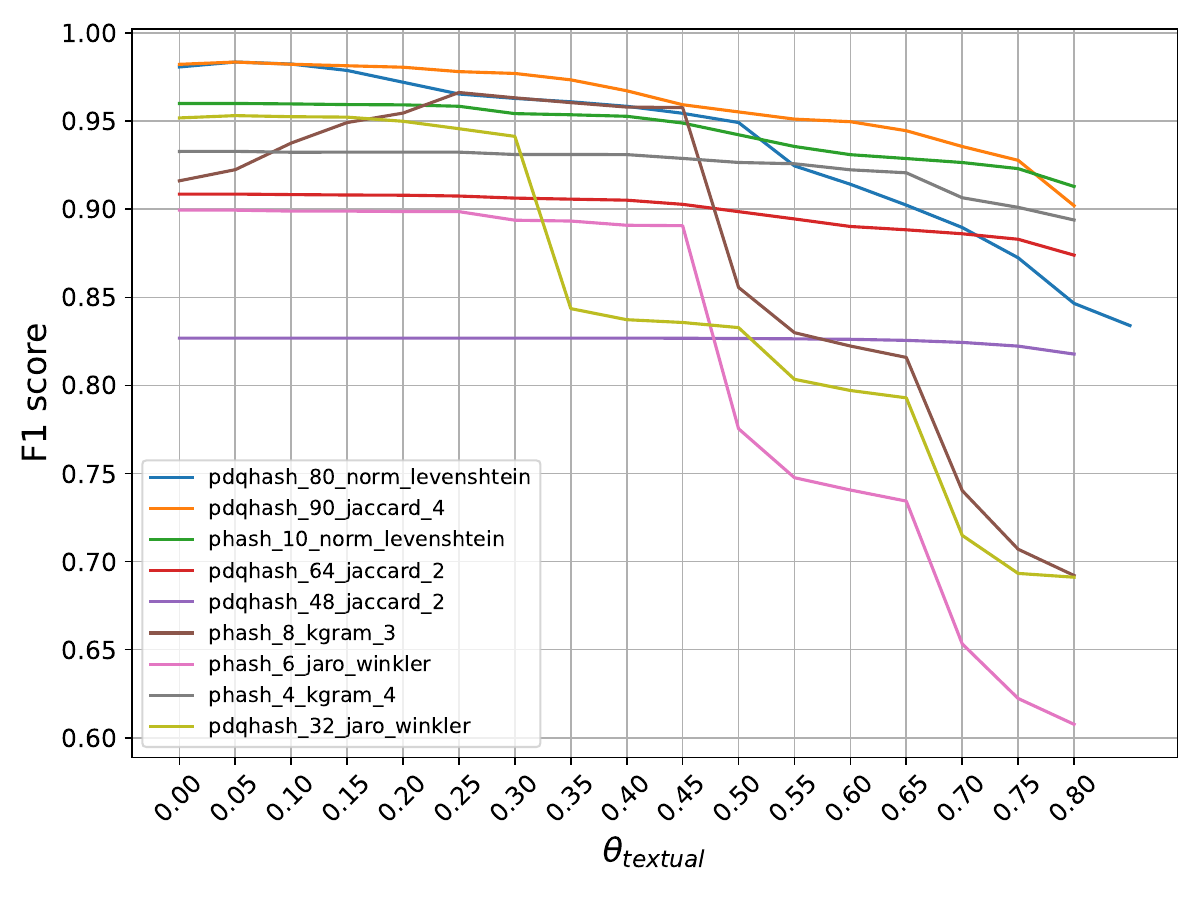}
 \caption{F1 score of different grid search components.}
 \label{fig:results_gridsearch}
\end{figure}

\descr{Grid search setup for $\theta_{visual}$ and $\theta_{textual}$ to determine the best operating values for \systemname.}
To identify the best set of components for \systemname, we experimentally determine the best hashing method, the corresponding distance threshold for that hash, similarity metric for the OCR label, and accordingly the similarity threshold for comparing $label_{query}$ and $label_{match}$.
We perform a grid search over these four different components of \systemname, scoring the combinations of the components by their corresponding F1-score, evaluated on ~\visualgroundtruth.
We present the results of the grid-search experiment in Figure~\ref{fig:results_gridsearch}.
For space reasons, we only present the best performing text similarity method for each $\theta_{visual}$ and the two hashing methods.
We analyze how the F1 score of the embeddings with $\theta_{visual}$ change as we increase the $\theta_{textual}$.

We can see that the combination of PDQHash with $\theta_{visual}$ of 90, and the OCR component using Jaccard similarity ($ngram = 4$) with $\theta_{textual}$ of 0.05 produces the best F1 score of 0.980.
This setting yields a precision of 0.990 and recall score of 0.979.
The next best performing metrics are PDQHash with $\theta_{visual}$ of 80 and pHash with $\theta_{visual}$ of 10 using normalized levenshtein similarities.
We observe the immediate returns of expanding the $\theta_{visual}$ to the maximum bound to retrieve as many relevant results as possible, without compromising on the false positives.
We will further evaluate this configuration with previous state of the art image detection methods in Section~\ref{sec:comparison_baseline}.
We also note that we can improve the performance of pHash embedding by using a wider $\theta_{visual}$ of 10, compared to the thresholds of 6 and 8 used in the prior works. 

In the rest of the paper, we set \systemname to use an image embedding of \textbf{PDQHash} with a $\theta_{visual}$ of 90 and an OCR post-processing component of Jaccard similarity ($ngram = 4$) with $\theta_{textual}$ of 0.05. 

\section{Evaluation}
In this section, we first compare \systemname with various other image similarity systems as baselines, showing that \systemname is the best performing approach for content moderation.
Next, we use \systemname to identify misleading images during the 2020 US Presidential Election in the wild, and compare its effectiveness with the existing moderation system of Twitter.
Finally, we perform a qualitative characterization of the misleading images by analyzing the images through the lens of Twitter's platform policies in detail.

\subsection{\systemname vs. baselines}
\label{sec:comparison_baseline}
We compare the performance of \systemname against various image similarity systems as a baseline for evaluation.
We use four different families of image similarity systems for comparison.
First, we compare \systemname against semantic embeddings that use Deep Neural Networks (DNNs) based methods. 
These include features extracted from final layers of DNN architectures specializing in image classification such as Inception-v3~\cite{szegedy2016rethinking} and ResNet~\cite{he2016deep}. 
The features extracted from these methods have been shown to be useful for visual search and similarity tasks~\cite{ahmed2019content,sklan2015toward}.
The next family of methods that we compare \systemname against is image descriptors, which can be considered as ``deeper syntactic hashes''~\cite{davis2019open}.
This set of methods is popular in computer vision tasks such as object recognition, scene recognition, and work by extracting a large number of \emph{keypoints} in an image.
Matatov et al.~\cite{matatov2018dejavu} use ORB descriptors for identifying the spread of visual misinformation to assist journalists.
We use ORB~\cite{rublee2011orb}, SIFT~\cite{lowe1999object}, and DAISY~\cite{tola2009daisy} as image descriptors.
Another family of method we compare \systemname with is the list of prior works that have used perceptual hashing algorithms for identifying images spreading misinformation.
These include Phash and PDQHash with multiple distance thresholds~\cite{javed2020first,kazemi2021tiplines,wang2020understanding,zannettou2018origins}.

Finally, we evaluate the performance of \systemname with multi-modal embeddings.
While there is a rich body of work and systems existing in the vision-language literature, the majority of these works are trained with an objective of explaining the visual concept contained in the image~\cite{lin2014microsoft}.
On the other hand, OCR texts occur as semantic context to the image itself and can be much longer, noisy, and nuanced.
Thus it is not possible to directly adapt models such as Universal Image Text Representation Learning (UNITER)~\cite{chen2020uniter}, or Visual-BERT~\cite{li2019visualbert} that have been trained on vision-language tasks for our problem, in the same way adapting Inception-v3 or ResNet is possible.
However, to assess the potential of a multi-modal embedding that leverages text from an OCR engine, or text from the accompanying tweet with DNN based features, we experiment with ``concatenated'' vision-language embeddings.
First, we concatenate the embeddings obtained from Inception-V3 (a 2048 dimensional vector) and Sentence-BERT~\cite{reimers2019sentence} (a 768 dimensional vector) to create a joint embedding from the two modalities.

We experimented with two different techniques for unifying the embeddings: i)~concatenating, and ii)~stacking, further reporting the best results.
While much more sophisticated mechanisms of combining embeddings such as attention mechanisms exist, the dataset size of \visualgroundtruth limits us in evaluating methods that combine embeddings or modalities. 
Fully leveraging multimodal embeddings requires an end-to-end deep learning training setup, and accordingly large scale datasets designed for the problem of soft-moderation, which currently does not exist.

In the same way, we also combine textual and visual embeddings using the CLIP model from Open AI~\cite{radford2021learning}.
CLIP consists of a text and an image encoder which encodes textual and visual information into a multimodal embedding space by using contrastive learning.
The model is trained on various text/images pairs from Web sources, and has demonstrated impressive zero-shot capabilities for classification purposes and have been used to detect hateful content on social media~\cite{gonzalez2023understanding,qu2023evolution}.
Alongside the OCR text, we also experiment with using the tweet text as part of the multi-modal embedding to evaluate if using the tweet text as further context can improve model performance for image retrieval.
To this end, we concatenate the visual information encoded through the image encoder with three different variants of textual information: i)~tweet text, ii)~tweet text + OCR text, and iii)~OCR text.
After extracting the joint embeddings using CLIP, we further normalize the multi-modal embeddings for evaluation.

For the neural network based image embeddings (Inception-v3, ResNet-50, ResNet-101, and ResNet-152) we experiment with multiple resolutions of input (1x, 2x, etc.), reporting the best performing F1 score for each method.
Similarly, for the neural network based image embeddings and the multi-modal embeddings, we experiment with multiple similarity thresholds ($0,0.05, 0.1, 0.15, \dots, 0.75, 0.8$) and report the best performing score for each method (except for the methods that use perceptual hashes with pre-determined similarity thresholds).
For the image descriptors, we experiment with a different number of features (30, 60, 90, and 120) and report the best-performing F1 score for each method.
We compare these methods on~\visualgroundtruth. 

\begin{table}[]
    \center
\scalebox{0.8}{
\begin{tabular}{|c|c|c|c|c|}
\hline
\textbf{Method} & \textbf{Prec.} & \textbf{Rec.}    & \textbf{F1}  & \textbf{Runtime} \\ \hline
Inception v3~\cite{szegedy2016rethinking}    & 0.679 & 0.878 & 0.766   & 0.031s  \\ \hline
ResNet-50~\cite{he2016deep}       & 0.518             & 0.938            & 0.667 & 0.027s     \\ \hline
ResNet-101~\cite{he2016deep}      & 0.617             & 0.932             & 0.742  & 0.030s    \\ \hline
ResNet-152~\cite{he2016deep}      & 0.440              & 0.962             & 0.604  & 0.034s     \\ \Xhline{3\arrayrulewidth}
ORB descriptors~\cite{rublee2011orb}             & 0.491             & 0.535            & 0.512  & 0.040s   \\ \hline
SIFT descriptors~\cite{lowe1999object}           & 0.935           & 0.0136           & 0.026  & 0.253s   \\ \hline
DAISY descriptors~\cite{tola2009daisy}      & 0.484           & 0.431             & 0.456 & 0.257s    \\ \Xhline{3\arrayrulewidth}
    PDQHash (thr. 32)~\cite{kazemi2021tiplines}      & 0.992           & 0.798           & 0.885  & 0.020s    \\ \hline
    PDQHash (thr. 40)~\cite{javed2020first}  & 0.975             & 0.838            & 0.901   &  0.020s  \\ \hline
    Phash (thr. 6)~\cite{wang2020understanding}           & 0.995             & 0.596             & 0.746    & 0.017s \\ \hline
    Phash (thr. 8)~\cite{zannettou2018origins}           & 0.991             & 0.707             & 0.826  & 0.017s \\ \Xhline{3\arrayrulewidth}
    Inception v3~\cite{szegedy2016rethinking} $+$ & & & & \\ 
SentenceBERT~\cite{reimers2019sentence} & 0.711 & 0.972 & 0.820 & 0.076s  \\ \hline
CLIP (tweet text)~\cite{radford2021learning} & 0.814  & 0.787 & 0.800 & 0.149s \\ \hline
CLIP (tweet text + OCR)~\cite{radford2021learning} & 0.862 & 0.810 & 0.835 & 0.149s \\ \hline 
CLIP (OCR)~\cite{radford2021learning} & 0.883  & 0.828 & 0.855 & 0.149s \\ \hline \Xhline{3\arrayrulewidth}
\textbf{\systemname}        & \textbf{0.990}            & \textbf{0.979}            & \textbf{0.980}  & 0.223s     \\ \hline
\end{tabular}
}
\caption{Comparison of \systemname with baselines.}
\label{tab:baselinecomparisons}
\vspace{-5mm}
\end{table}

The results of the comparison of \systemname with the baselines are presented in Table~\ref{tab:baselinecomparisons}.
Image descriptor methods perform poorly when used for soft moderation. 
This is because these approaches are designed for higher level tasks like object and scene recognition.
Deep neural network approaches work well in identifying the subjects of moderated images (e.g., Donald Trump or Joe Biden) and therefore report a high recall.
At the same time, however, their precision is low, because they flag any image containing the same subjects as similar.
We find that leveraging multimodal embeddings slightly improves the performance over DNN based methods, but still has a very low precision (0.711) for Inception-v3+SentenceBERT.
In the same way, using concatenated embeddings from Open AI's CLIP significantly increases the precision over single-modality DNN methods, but has a very low recall (0.828).
Among the three different variants of CLIP embeddings leveraging different channels of modalities, we find that encoding the most simplest channel, i.e. CLIP (OCR) has the best performance. 
Surprisingly, encoding tweet text alongside the OCR text, i.e., both CLIP (tweet text) and CLIP (tweet text + OCR) does not improve the performance of the CLIP model.
This can be attributed to how state-of-the art multi-modal embeddings like CLIP are designed to match image and caption pairs, and therefore fail to capture the nuances of text and media co-usage in social media.
Perceptual hashing approaches, on the other hand, identify visually similar images, and therefore report a high precision.
At the same time, the need to exclude images that are visually similar but contain text that is contextually different forces these approaches to use low similarity thresholds, which limits their recall.

\systemname overcomes the limitations of all three types of approaches.
The use of perceptual hashing allows our approach to have a better precision than neural network based ones.
At the same time, the use of OCR to determine the context of an image allows \systemname to operate at a higher threshold than existing systems based on perceptual hashing, addressing the low recall reported by these methods.
While \systemname's precision is slightly lower than the best-performing perceptual hash method (0.990 vs. 0.995), its recall is the highest among all tested approaches.
As a result of this, \systemname reports the best F1-score among the tested approaches, balancing false negatives and false positives better than previous work.

We also compare the runtime of different systems in Table~\ref{tab:baselinecomparisons}.
For each method, we report the combined time of generating the image embeddings and indexing those embeddings on Milvus, averaged over 5 independent runs on identical system load.
The time for retrieving visually similar images from Milvus is not considered as Milvus optimizes the retrieval time across all embeddings of different sizes (average of 0.27 seconds) for all of the systems.
We can observe that the runtime of \systemname is around 8 times slower than DNN based methods, and about 10 times slower than peceptual hashing based methods.
Computing the OCR label of an image is the most time consuming operation of \systemname compared with other methods as it takes an average of 0.223 seconds per image.
While this is a large overhead incurred by \systemname, we need to keep in mind that its OCR component is only triggered when a similar image to a seed one is identified within the threshold of $\theta_{visual}$, which only occurs for 0.973\% of the images in \visualgroundtruth.
When there is no match, the time overhead of \systemname is the same as PDQHash.
Therefore, we argue that this slowdown of 0.2 seconds every 100 images on average is an acceptable tradeoff, allowing \systemname to improve its F1-score by 8\% over the second best performing algorithm and allowing for more comprehensive soft moderation.

\begin{figure}{}{}
 \centering
   \includegraphics[width=\linewidth]{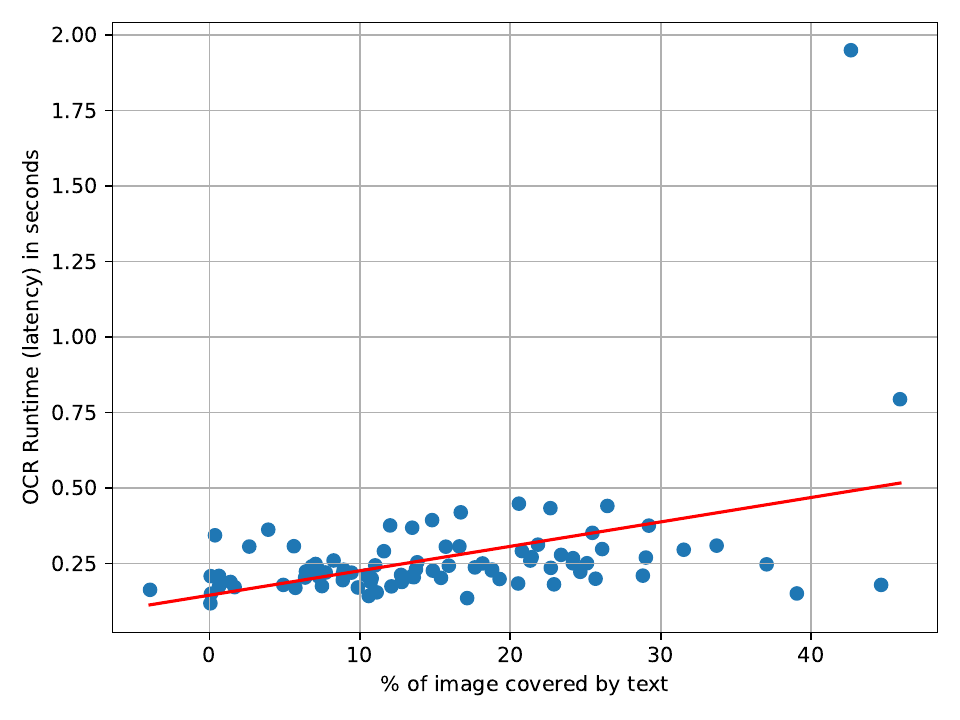}
 \caption{Latency of OCR by changing the percentage of text covering images in~\visualgroundtruth.}
 \label{fig:ocr_latency_text}
\end{figure}

To further check the impact of the OCR component of \systemname, we examine how the latency of OCR changes with increasing amount of text in the images.
We characterize the amount of text by using the percentage of image area covered by it.
Figure~\ref{fig:ocr_latency_text} shows a scatter plot of the OCR runtime against the percentage of image covered by text in~\visualgroundtruth.
The Pearson's correlation coefficient ($r$) between the percentage of image covered by text and runtime is 0.401, suggesting a moderate positive correlation between runtime and amount of text contained in an image.
This moderate correlation indicates that the fraction of image covered by text plays a role in the OCR runtime.

\subsection{Detecting Visual Misleading Information on Twitter using \systemname}
\label{section:detection_wild}

In this section, we evaluate \systemname in the wild to further identify images that, while spreading false information, were not moderated by Twitter.
This showcases the utility of our approach while also highlighting the serious need for scalable and automated techniques to identify visually misleading information.
We first build a set of seed images spreading misleading information, using the 2020 US Presidential Election as a case study.
We then apply \systemname to find more images that are similar to the seeds that \emph{should have been} moderated but were not.
Finally, we perform a thorough qualitative analysis of the types of images detected, providing a characterization of image-based misinformation in the context of the 2020 US Presidential Election and how they align with Twitter's platform policies.

\subsubsection{Identifying and Filtering 2020 US Presidential Election misleading images}
\label{sec:filtering}

In this section, we first describe how we select a dataset of images that received soft moderation from Twitter in the context of the 2020 US Presidential Election.
We then present the results reported by \systemname when looking for similar images in the wild, and compare these results with the coverage of the moderation applied by Twitter.

\descr{Building a seed set of misleading images.}
To build a seed set of misleading images contained in tweets that were moderated by Twitter, we use datasets ~\done and ~\dthree. 
Following the methodology of~\cite{zannettou2021won}, we check the moderation status of all tweets in \done during late 2022 and find 1,133 tweets that were moderated by Twitter.
Unfortunately, ~\dthree is too large (6.9M tweets) for us to check the moderation status of each tweet, given the limitations imposed by the Twitter API. 
Therefore, we only check for the tweets containing an image that occurred more than 5 times in ~\dthree, based on their PDQhash, resulting in 542,700 candidate tweets.
From these, we identified a further 3,134 tweets with images that were moderated.
Combining the moderated tweets from these two sources, we end up with a final seed set of 4,267 images.

An issue that we face at this stage is that our dataset contains tweets that were moderated by Twitter, but we do not know if the moderation decision was made based on the image included in the tweet or based on the text.
In fact, it is rather common for social media users to include generic images (e.g., memes or GIFs) in their posts
that do not contain false information themselves.
Two examples of these types of tweets are presented in Figure~\ref{fig:contextually_misleading} (in the Appendix).
Neither image conveys false information, but the text is clearly supporting false claims of election fraud. 
If we kept these images in our seed dataset, \systemname would flag many unrelated tweets that just happen to include them, greatly reducing its effectiveness.

\begin{figure}[t]
     \centering
     \begin{subfigure}[b]{0.49\columnwidth}
         \centering
         \includegraphics[width=\textwidth]{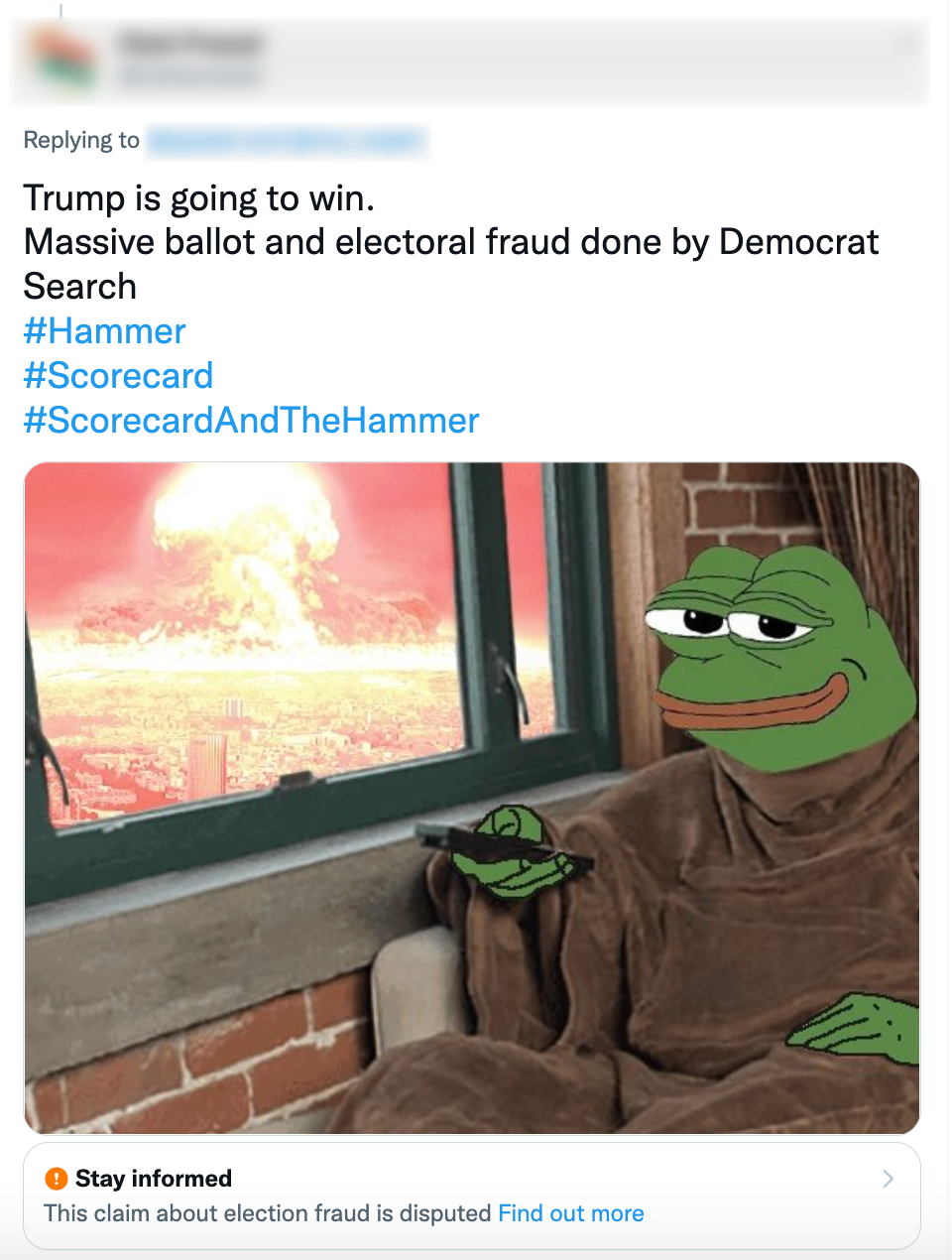}
         \caption{Comfy Pepe meme used as media on a tweet.}
         \label{fig:contextual_1}
     \end{subfigure}
       \hfill 
     \begin{subfigure}[b]{0.49\columnwidth}
         \centering
         \includegraphics[width=\textwidth]{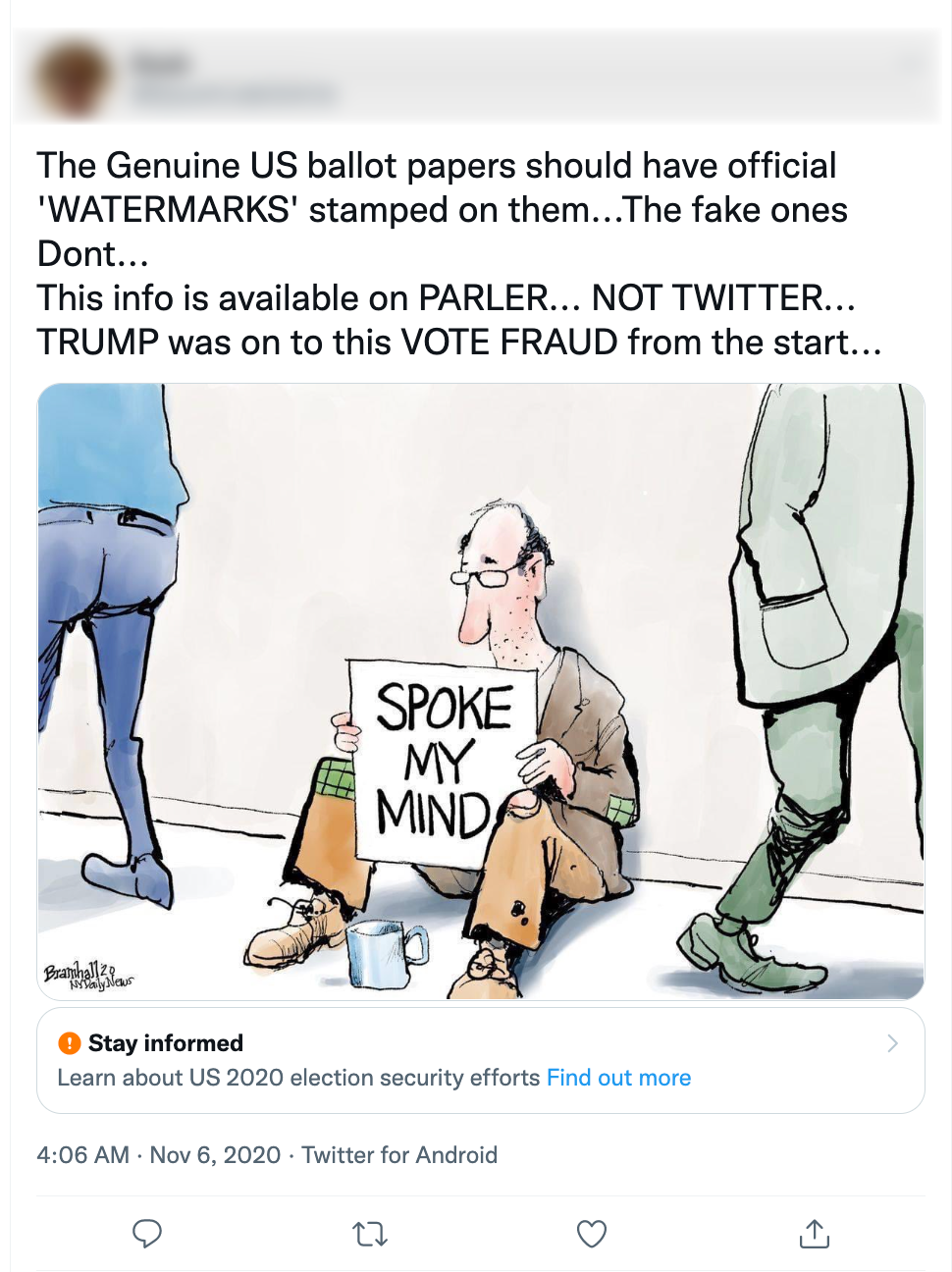}
         \caption{A cartoon illustration used as media on a tweet.}
         \label{fig:contextual_2}
     \end{subfigure}
        \caption{Moderated images irrelevant to the 2020 election.}
        \label{fig:contextually_misleading}
\end{figure}

To avoid this issue, we manually annotate the 4,267 tweets with images that were moderated by Twitter, discarding those not directly related to false claims about the 2020 US Presidential Election.
We first sample 200 images from the 4,267 seed images and have two annotators independently label them as being relevant or not.
The annotators then discuss their results, finding that 186 out of 200 images were relevant with nearly perfect agreement (Cohen's Kappa score of $\kappa$ = 0.9177~\cite{mchugh2012interrater}).
This indicates that an image's relevance to the 2020 US Election is clearly understood between the two annotators.
We then split the remaining 4,067 seed images between them and wind up with a total of 959 relevant images.

\descr{Detecting misinformation images using \systemname.}
Finally, we use \systemname to retrieve more tweets from the index that contain similar images to the seed misleading images.
To this end, we initially retrieve 50,439 images visually similar to the 959 seed misleading images, out of which 40,244 are further filtered to be contextually similar to the query images, and hence candidates for moderation.

To check for the accuracy of these moderation candidates identified by \systemname, we perform a false positive and false negative analysis. 
First, we randomly sample 50 \emph{query images} from the 959 seed misleading images.
Since the 50 \emph{query images} are already manually verified to contain misleading images related to the 2020 US Presidential Election, we can assume that visually and contextually similar images to the images will be misleading images as well.
We retrieve and annotate all the images from our index visually similar to the 50 \emph{query images} following the similar methodology used to build \visualgroundtruth in Section~\ref{sec:system_evaluation}, resulting in a reference set of 10,387 images. 
Upon querying \systemname with the same 50 \emph{query images}, it retrieves 10,172 images both visually and contextually similar to the \emph{query images} as moderation candidates.
We find that 96 of these detected images are both visually and contextually dissimilar to the corresponding \emph{query images}, resulting in a false detection rate of 0.99\%.
This is due to the well-known limitations of perceptual hashing algorithms~\cite{zannettou2018origins}, for example when dealing with images where a solid color background dominates.
On the other hand, we find that \systemname has a false negative rate of 2.06\% as it could only succesfully detect 10,172 images out of the possible 10,387 images.
Based on these results, \systemname performs a lot better than Twitter's internal soft moderation approach, which moderated only 521 of those 10,387 images (94.98\% false negatives).

Finally, we check if the candidate tweets containing the images identified by \systemname were also moderated by Twitter.
Following the methodology in~\cite{zannettou2021won}, we extract the relevant metadata related to soft moderation interventions for the tweet.
Out of the 40,244 tweets checked, we find that only 2,950 tweets received soft moderation interventions by Twitter.
We were unable to check the intervention-related metadata for 2,479 tweets as they were inaccessible during the time we conducted the study.  
This analysis indicates that Twitter's soft moderation approach is inadequate, with a large proportion of tweets (92.66\%) containing misleading images left without moderation, and it showcases the utility of automated approaches like \systemname.

\subsubsection{Characterizing misleading images about the 2020 Presidential Election on Twitter}
\label{sec:category}

In this section, we perform an in-depth analysis of the images identified by \systemname.
To aid human annotation, we first group visually similar images into \emph{Image Stories}.
We then code the extracted \emph{Image Stories} according to Twitter's platform content policies~\cite{twitter_manipulated_media,twitter_civic_info}, with the goal of understanding how the detection performed by \systemname aligns with the platform's existing moderation guidelines.
Finally, we investigate whether misleading images violating different types of content policies get moderated differently by Twitter.

\descr{Aggregating misleading images into \emph{Image Stories}.}
To help manually code the images detected by \systemname, we aggregate them into \emph{Image Stories}.
We define an ``Image story'' as a set of similar images that convey the same misleading message in the same context.
For example, the three images included in three different tweets in Figure~\ref{fig:tweet_moderation_examples_img} are visually similar, and should be grouped together.
To this end, we follow a grouping approach similar to the one adopted by Zannettou et al. when studying image memes~\cite{zannettou2018origins}.

First, we apply clustering to the PDQhash embeddings of the images detected by \systemname to group visually similar images by using the DBSCAN clustering algorithm~\cite{ester1996density}.
To use the DBSCAN clustering algorithm, we need to select parameters for two parts of the process: i)~distance threshold for assessing similarity, and ii)~minimum number of elements in a cluster.
Through the experimental validation performed in Section~\ref{sec:system_evaluation}, we have already established that using PDQhash as the syntactic embedding, with a $\theta_{visual}$ of \textit{90} is best suited for identifying visually similar images.
We use this empirically informed distance threshold as the Hamming distance difference threshold for the first clustering parameter and set the minimum number of elements in a cluster to be 1, to allow for misleading images that occur in isolation, without visual variants.  
Upon using the DBSCAN algorithm to cluster the 40,244 images detected by \systemname, we obtain 258 clusters of images.
These image clusters represent the misleading images aggregated into visually similar claims conveying the same misleading message.

\descr{Analyzing misleading images through the lens of Twitter's Platform Policies.}
We present our codebook, which guides our annotation process for characterizing \emph{Image Stories} used during the 2020 Presidential Election.
We develop the codebook such that it aligns with the platform policies of Twitter~\cite{twitter_manipulated_media,twitter_civic_info}.
We use two different policies outlined in Twitter's \emph{Platform Integrity and Authenticity} resources: i)~Synthetic and manipulated media policy~\cite{twitter_manipulated_media} and ii)~Civic Integrity Policy~\cite{twitter_civic_info}.
Each policy contains multiple categories of violations, and each category is broken down into multiple rules.
The Civic Integrity Policy outlines how ``Twitter should not be used for the purpose of manipulating or interfering in elections or other civic processes such as refereed, censuses.''
We include 3 different categories from the Civic Integrity Policy out of the four available, which we discuss in further detail.
The fourth category, ``False or misleading affiliation'' is related to the use of fake and parody accounts, and thus not related to misleading images.
Similarly, we include the ``Synthetic and Manipulated Media Policy'' as this policy covers usage of media on Twitter, which is the major modality of content in our study.
The details of the individual rules and categories are listed in Appendix~\ref{sec:appendix_policy}.

To begin the process, we first randomly sample 100 \emph{Image Stories} out of the 258 extracted to guide us in developing the codebook.
First, we take each of the 16 different rules of Twitter's Civic Integrity Policy~\cite{twitter_civic_info}, divided into four major categories as the draft of our initial codebook.
Two annotators independently code the sampled images, considering additional contextual information learned from fact-checking articles associated with them.
We discuss these codes, repeating the process three times until the final codebook reached a point where further iterations would not improve it.
Note that we use the 16 different rules to help us identify the closest violation of Twitter's platform policy by the images while annotating the \emph{Image stories} with the granularity corresponding to the four categories associated with the rules.
We reach a nearly perfect agreement (Cohen's Kappa score of $\kappa$ = 0.9~\cite{mchugh2012interrater}).

Positioning the extracted \emph{Image Stories} alongside the platform policies of Twitter helps us re-evaluate the content moderation decision based on Twitter implementations and policies.
We present an example image violating each of the four categories discussed in Figure~\ref{fig:combined_categories} (in the Appendix).
In the rest of this section, we describe the four categories in our codebook in detail, along with their definitions and our evaluation for matching images with the corresponding category.

\begin{figure}[t]
    \centering

    \begin{subfigure}[b]{0.49\columnwidth}
        \centering
        \includegraphics[width=\textwidth]{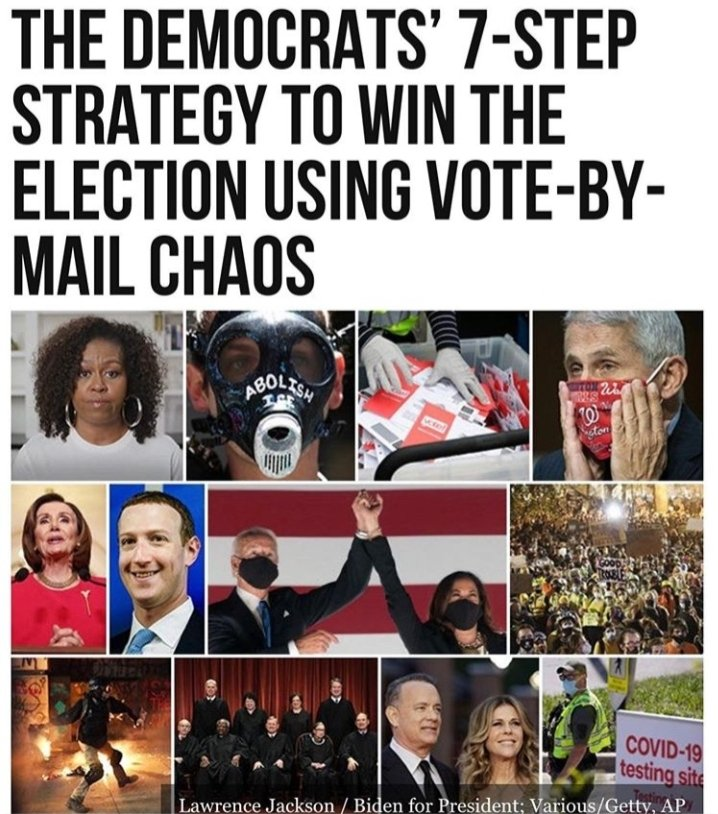}
        \caption{Participation in civic processes.}
        \label{fig:content_participation}
    \end{subfigure}
    \hfill
    \begin{subfigure}[b]{0.49\columnwidth}
        \centering
        \includegraphics[width=\textwidth]{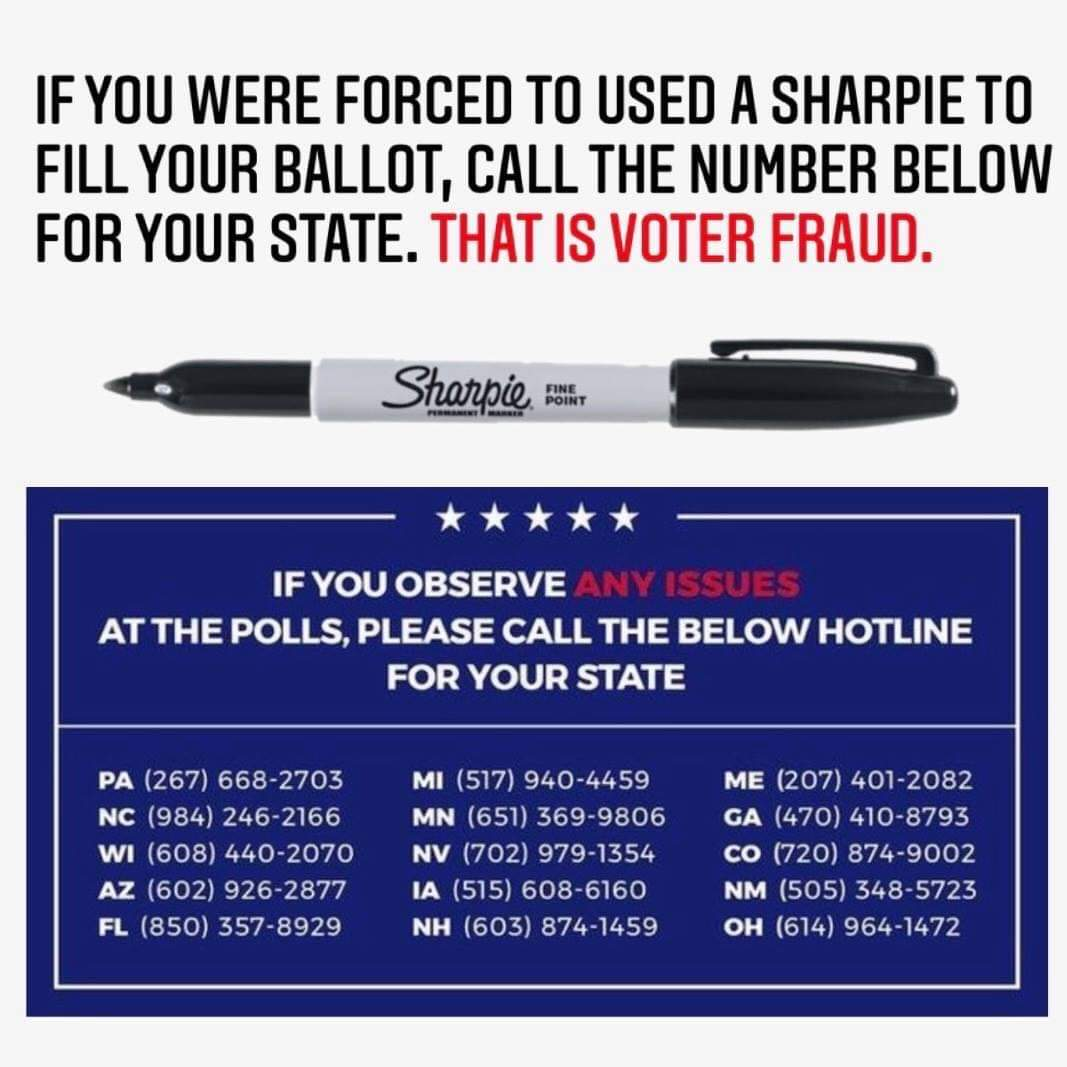}
        \caption{Intimidate people from civic processes.}
        \label{fig:content_intimidation}
    \end{subfigure}

    \begin{subfigure}[b]{0.49\columnwidth}
        \centering
        \includegraphics[width=\textwidth]{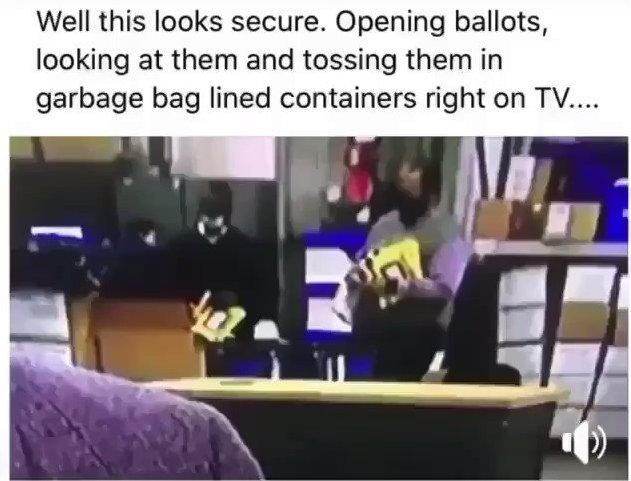}
        \caption{Outcomes of civic processes.}
        \label{fig:content_outcomes}
    \end{subfigure}
    \hfill
    \begin{subfigure}[b]{0.49\columnwidth}
        \centering
        \includegraphics[width=\textwidth]{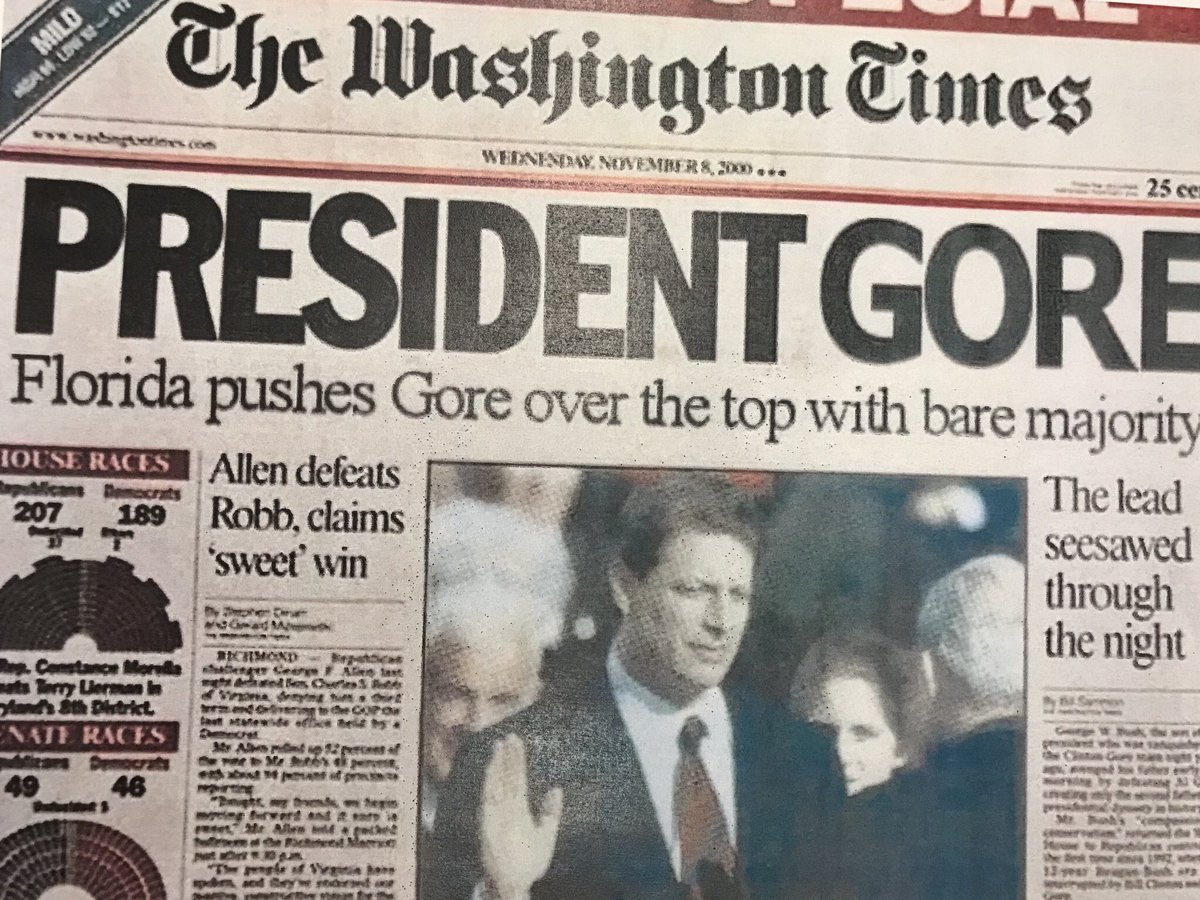}
        \caption{Synthetic and Manipulated Media.}
        \label{fig:content_synthetic}
    \end{subfigure}

    \caption{Example images violating four major categories of Twitter's Civic Integrity Policy.}
    \label{fig:combined_categories}
\end{figure}

1. \textbf{\textit{Misleading information about how to participate in civic processes}}:
This category refers to images that mislead people about ``election participation procedures and requirements, cause confusion about officials, or discuss threats to voting locations.''
Note that we did not find any images related to threats at voting locations in the extracted \emph{image stories}.
For election participation procedures and requirements, misleading images aim to spread false claims about the irregularity of the criteria by which votes are being cast.
For example, dead people are casting votes; votes are cast multiple times; votes are being cast after the closing of polls.
Misleading images may also claim that mail-in ballots are insecure, illegal, and a source of voter fraud.
Figure~\ref{fig:content_participation} aims to intimidate voters from voting-by-mail, casting doubts over the security of mail-in ballots despite mail-in ballots proven to be safe to use~\cite{kiely_2020}. 

2. \textbf{\textit{Misleading information intended to intimidate people from civic processes}}:
This category contains images that mislead people about ``how votes are being counted (or not counted), problems with ballot equipment, disruption at voting locations, and the closing of polls''.
Misleading images in this category aim to make claims about legitimately cast ballots getting invalidated, malfunction of voting machines, switching of votes between candidates, and deleting of votes on the machines.
We find images spreading misleading claims about disruptions at voting locations like poll workers filling ballots, poll workers forcing voters to use Sharpies, and ballot observers not being allowed to observe the counting process.
The example in Figure~\ref{fig:content_intimidation} is spreading false claims about ballots using Sharpies in Arizona being invalidated and poll workers forcing people to use Sharpies~\cite{staff_2020}.

3. \textbf{\textit{Misleading information about the outcomes of civic processes}}:
This category refers to the images that mislead people about ``election rigging, ballot tampering, vote tallying, declaration of premature victory, casting doubt on the outcome of civic processes, calling for interference with the implementation of election results, and undermining public confidence in the methods and results of the election.''
Misleading images in this category aim to make claims about the illegal processing and handling of ballots, which include alteration, manipulation, destruction, forging, counterfeiting, stealing, and pre-filling ballots.
We also find false claims about voter registration and turnout, and visual and statistical anomalies in patterns of vote counting.
By casting doubt on the outcome of civic processes, misleading information aims to question the integrity of the election process, discredit different stages of election processes, and call for interference with the implementation of election results using slogans such as ``Stop the Steal,'' ``Fraud.''
The misleading image in Figure~\ref{fig:content_outcomes} casts doubt over bipartisan vote tallying process, suggesting that Trump ballots were discarded in garbage bags~\cite{kiely_2020}.

4. \textbf{\textit{Synthetic and Manipulated Media}}:
This category refers to images that are ``significantly and deceptively altered, manipulated, or fabricated, and images shared with deliberate intent to deceive people''.
For this category, included images are significantly and deceptively altered, manipulated, or fabricated.
On the other hand, images shared with malicious intent include out-of-context tweets sharing the media.
For example, adding quotations shared in the past that are taken out of context to spread election misinformation or images from the past that are re-used in the present context, like discarded ballots from the 2018 election being repurposed to say ballots are being tampered with in the 2020 election.
Figure~\ref{fig:content_synthetic} is a doctored image in connection to the ``Bush-Gore Florida recount'' during the presidential race in 2000~\cite{alvarado_2020}, shared by a Trump campaign spokesman.

We assign each \emph{Image Story} to \emph{one} category.
There are some observations from our annotation process:
We notice that the use of memes is common in spreading misleading information.
They are designed for people to add text to them to make a variation.
During the annotation process, we first assess the overlay text on the memes to see if it falls into any of the categories other than the fourth: Synthetic and Manipulated Media. 
If it does not, we classify it as an image altered to be shared with malicious intent.

We also find that a large portion of images are used out of context and could potentially be categorized into the fourth category: Synthetic and Manipulated Media.
However, out-of-context information is widely used to push narratives related to the other three themes.
Thus, we take precedence of annotating the images into the first three categories if it suits any.
Otherwise, it is annotated as the fourth category.
We annotate these images according to their primary content to expand and deepen the qualitative analysis of these moderated images.
For example, a screenshot of a video clip in which ballot workers are transcribing ballots is used out of context and spread with the misleading content of ``ballot stuffing.''
In this example, the medium of spreading the misleading image is taking it out of context. However, the primary content being spread is related to vote counting and ballot tampering, which falls under the Twitter Civic Integrity Policy.
Thus, we annotate it as Category Three: Misleading information about the outcomes of civic processes.

\descr{Moderation of misleading images by violation category.}
After associating the misleading images with the corresponding platform policy violations, we want to understand whether Twitter is more likely to moderate images violating some of them.
 Table~\ref{tab:category_breakdown} reports the breakdown of the rate of moderation by Twitter of images belonging to the different categories of violations, among the tweets identified by \systemname.
A high-level overview of the breakdown by category suggests that Twitter moderated specific types of misleading \emph{image stories} more aggressively, while the moderation rate on other categories appears to be laxer.
Misleading \emph{image stories} of the categories ``Intimidation from Civic Processes'' and ``Participation in Civic Processes'' are moderated the most, with 5.96\%, and 4.16\% of the images getting moderated.
Surprisingly, the most popular category of content violation among \emph{image Stories}, ``Misleading information about outcomes of civic processes'' has the lowest moderation rate of 1.77\%.
Finally, images violating the category ``Synthetic and Manipulated Media'' have the moderation rate of 2.76\%.
While we can make overall observations that different types of platform policy violations might have been disproportionately moderated from Twitter's end, we find concerning results of overall moderation rate across the categories being very low.

\begin{table}[t]
\centering
\scriptsize
    \setlength{\tabcolsep}{4pt}
\begin{tabular}{lcc}
\toprule
\textbf{Category} & \textbf{\# \emph{Image stories}} & \textbf{Moderation \%} \\ \midrule
Participation in Civic Processes               & 57    &  4.16\%                  \\
Intimidation from Civic Processes              & 78  & 5.96\%                    \\
Outcomes of Civic Processes              & 81 & 1.77\%                         \\
Synthetic and Manipulated Media              & 42 & 2.76\%                         \\ \bottomrule
\end{tabular}
\caption{Moderation rate of images breakdown by category.}
\label{tab:category_breakdown}
\end{table}

\noindent

\section{Related Work}

\descr{Soft Moderation Approaches by Social Media.}
Social media platforms namely Facebook, and Twitter have increasingly shifted their approach towards warning labels as a tool for content moderation, and additionally, provide surrounding  context to the users when interacting with potentially false information.
These platforms have applied warning labels on false information shared ranging from Covid-19 pandemic~\cite{rosen_2021,culliford_2021,romo_2020,culliford_2020}, 2020 US Presidential Elections~\cite{abril_2020,graham_rodriguez_2020,bond_2020,conger_2020} to climate denial misinformation~\cite{connellan_2021,calma_2021}.
Twitter has reported that approximately 74\% of the tweet viewership happened after the warning labels were applied to the tweets and, warning labels were effective in decreasing users quoting the misleading tweets by an estimated 29\%. 
Savvas Zannettou~\cite{zannettou2021won} performed empirical analysis on a sample of 2,244 tweets with warning labels related to the 2020 US Presidential Elections.
Paudel et al. proposed \textsc{Lambretta}, a system that aims to improve soft moderation of textual content on Twitter by leveraging Learning to Rank~\cite{paudel2022lambretta}.
To the best of our knowledge, \systemname is the first end-to-end approach to perform soft-moderation of images social media that has been proposed by the research community.

\descr{Applications of Perceptual Hashing in Computer Security.}
Perceptual hashing techniques have been widely used in computer security, leveraging them to detect phishing websites~\cite{kintis2017hiding}, scam websites~\cite{kharraz2018surveylance,miramirkhani2016dial},fraudulent services~\cite{nikiforakis2014stranger}, and to identify impersonators on social media~\cite{goga2015doppelganger}.
Perceptual hashes have also been used for content authentication~\cite{ahmed2010secure,wang2015visual}, and tamper detection on images~\cite{zhao2010perceptual,tang2008robust}. 
Alkhowaiter et al.~\cite{alkhowaiter2022evaluating} evaluated six types of perceptual hashes on their ability to detect image manipulation and transformation over two major social media platforms: Facebook and Twitter.
Historically, Microsoft developed PhotoDNA~\cite{ith2015microsoft}, a type of syntactic embedding to identify and report the distribution of child exploitation material.
Facebook currently uses PDQHash in the ``ThreatExchange'' platform\footnote{https://developers.facebook.com/docs/threat-exchange/} to share ``signals'' of harmful media on their platform.

\descr{Image-based misinformation in social networks.}
Several studies are focusing on developing automated detection methods to identify images containing misinformation, either manipulated images~\cite{abdali2021identifying,bayar2016deep,zhou2018learning}, or images that are taken out of context or misinterpreted on social media~\cite{abdelnabi2022open,aneja2021cosmos,dewan2017towards,jin2016novel,zlatkova2019fact}.
Most of the works on multi-modal misinformation tackle classifying or detecting a single image as misinformation, while very few works have focused on studying the spread, and diffusion of misleading images within and across social media.
The closest work to \systemname is a system called DejaVu~\cite{matatov2018dejavu}, which is designed to assist journalists in collaboratively addressing the spread of visual misinformation by using ORB descriptors~\cite{rublee2011orb} (another type of syntactic embedding) to encode the images, and FAISS~\cite{johnson2019billion} to index the images.
Additionally, other works study the spread of COVID-19 media through WhatsApp in Pakistan~\cite{javed2020first}, and other types of visual misinformation in WhatsApp~\cite{kazemi2021tiplines,reis2020can}.
On the other hand, works by ~\cite{ng2022coordinated,zannettou2019characterizing} study the usage of images in state-sponsored influence campaigns.
Wang et al.~\cite{wang2020understanding} analyzed the spread of \emph{Fauxography} images on social media, which are images that are presented in an incorrect or untruthful fashion, by using ground truth fact-checked images from fact-checked organization Snopes~\footnote{https://www.snopes.com/fact-check/}.
Similarly, Zannettou et al. used visual similarity (i.e., perceptual hashing) and images annotated by the website KnowYourMeme to study the evolution and diffusion of image memes posted on social media~\cite{zannettou2018origins}.

\section{Discussion and Conclusion}
\label{sec:discussion}

In this paper, we presented \systemname, a scalable system able to identify images that are candidates for soft moderation on Twitter.
\systemname overcomes the inability of perceptual hashing to discern the context of an image by incorporating OCR into the matching process.
Our results show that \systemname outperforms three types of image matching systems based on perceptual hashing, on image descriptors, and on deep neural networks.
With the highest F1-score among competitors, \systemname places itself as the state of the art in automated image-based soft moderation.

We believe that \systemname (which we make publicly available \footnote{https://github.com/idramalab/pixelmod}) will be an inspiring foundation for researchers and online platforms aiming to improve content moderation on social media.
In the rest of this section, we first discuss the ethical considerations of our work and design implications that online platforms should keep in mind when deploying \systemname.
We then discuss the limitations of our approach and some directions for future work.

\descr{Ethical Considerations.}
Our work only uses publicly available Twitter data that was collected the official API while it was still open to academic researchers, and we do not interact with users.
As such, this work is not considered human subject research by our institution.
We also preserve the privacy of Twitter users as we do not analyze any personally identifiable information (e.g., location data, account names) and blur the regions of example tweets used in the paper containing identifiable (meta)data.
Additionally, we take steps to blur the example images used in the paper unless i)~they are public figures, ii)~they are an illustration, iii)~they are stock images.
As any content moderation system, \systemname could be used for malicious purposes like censorship and surveillance.
We advocate that the system should be used with ethical principles in mind, following the \emph{respect for public interest} and \emph{beneficence} principles from the Menlo report~\cite{kenneally2012menlo}.

\descr{Design implications.}
Our experiments found that Twitter soft moderation misses most images that should be labeled.
This means that leveraging \systemname would help Twitter's moderators cover more false information on their platform.
However, when deploying \systemname, Twitter or other online platforms should take several aspects into consideration.
First, while \systemname's detection performance exceeds that of existing approaches, a false detection rate of 0.99\% may still be considered too high by online platforms to consider its adoption as a fully automated soft-moderation system.
We envision PixelMod to be used by platform moderators as a tool to identify a set of candidates for soft moderation with limited false positives, which then receive manual vetting.
The ultimate decision on whether to apply the moderation labels, however, should remain with the human operator.
This is not dissimilar from what online platforms are doing already, but our approach would allow them to obtain a more comprehensive view of misleading content on their network.
Additionally, \systemname could help addressing the main pain points of relying on human moderation, which are the latency in decision making and having limited moderator resources~\cite{arsht2018human,steiger2021psychological}.

Second, \systemname is an inherently reactive system: it requires a set of images already identified by the platform as misleading.
This process could be streamlined by using example images that have been fact-checked by dedicated organizations~\cite{wang2020understanding}.
When curating the set of seed images in Section~\ref{sec:filtering}, we found that querying images directly from a list of moderated tweets can be tricky, since all of the moderated images might not be of misleading nature.
In such cases, moderators using \systemname should take an additional step to ensure the query images are misleading in nature, rather than ambiguous images not related to the events being studied, to get the most relevant results as candidates for moderation.

Misleading images can spread across multiple social media platforms, propagating with different contexts and forms~\cite{hunt2020misinformation,wilson2020cross}.
Platforms can use \systemname as a tool alongside an industry shared database of known misleading image hashes for tracking the spread of misleading images on their service and across other online communities.
Platforms already use shared databases for tracking Child Sexual Abuse Material (through the National Center for Misleading and Exploited Children) and terrorism content (through the Global Internet Forum to Counter Terrorism)~\cite{,zoom_csam,google_csam,newsroom_2016}.
A tool like \systemname backed by a centralized database could easily be added to major social media platform's existing efforts to combat fraud and misinformation~\cite{verge_platforms}.

\descr{Runtime implications.}
As discussed in Section~\ref{sec:comparison_baseline}, the increased runtime overhead of \systemname is due to the OCR component, which only gets triggered for 0.973\% of the images in \visualgroundtruth.
This rate is 1.450\% among the 19.7M images used in our ‘in-the-wild’ evaluation in Section~\ref{sec:filtering}.
It is to note that factors like image resolution, background and font complexity could also impact the runtime of the underlying OCR engine.
The overhead reported is an upper bound on the runtime of \systemname and the average runtime of our system is in the same order as other baselines.
This occasional slowdown is a tradeoff we make for increased recall for \systemname, which allows us to achieve a much higher coverage of soft-moderation candidates than the ones achieved by Twitter.
However, \systemname can be adapted according to the content moderation budget of the platform.
The initial set of results retrieved by \systemname (visual matches to query images) can be sorted or filtered through metadata such as popularity of the account posting the images, or other metadata aligning with specific content moderation strategies of a platform, before passing through the contextual similarity component.

\descr{Applying \systemname to other platforms and topics.}
We could not test \systemname on other topics and platforms due to the lack of reliable datasets across platforms and topics.
First, we are not able to test our system on other social media platforms due to the lack of access to a uniform sample of posts (like the 1\% sample that forms our \dtwo dataset)
While datasets containing misleading images exist for other platforms like WhatsApp and Telegram, these are not suited for our evaluation since they only contain a single instance of labeled misleading images, making the visual similarity research process that is at the center of \systemname moot.
Even though Twitter applied warning labels on COVID-19 misinformation, these were unreliable and inconsistent~\cite{lange_2020,lyons_2020}, which we independently confirmed in our preliminary analysis.
Despite the limitations on evaluation settings, we expect \systemname to generalize well across multiple platforms and topics.
The only requirement for a platform to apply \systemname to a new campaign is a set of seed images that are known to be misleading, and the system should generalize well to other platforms and topics, as the underlying image embeddings are syntactic in nature, and do not incorporate any domain-specific metadata (e.g. number of retweets, number of likes available on tweets).

\descr{Limitations.}
Despite the promising performance of \systemname in identifying visually similar images at scale, the embedding used by \systemname (PDQHash) is vulnerable to adversarial manipulations~\cite{hao2021s,hu2022badhash}.
An adversary could modify images to have a PDQhash that is very far from the one of the corresponding seed image, generating a false detection by \systemname.
This is a serious risk, and future research should investigate defenses against these attacks.
Some potential avenues of defense are utilizing an ensemble of hashes: combining results from both pHash and PDQHash~\cite{hao2021s}, and adversarial training of embeddings~\cite{wang2021targeted}. 
Using an ensemble of hashes would force an attacker to jointly optimize the adversarial attacks against an ensemble of hashing methods opposed to a single method, thus increasing the operation cost on the end of adversaries.
Future works can also look at incorporating the contextual information (OCR text) contained in the images as part of generating the syntactic embeddings themselves, making it difficult for adversaries to modify the images without deviating from the contextual messaging of the images.
At the same time, the threat model here assumes centralized coordination by an adversary.
While this is within reach when dealing with state-sponsored disinformation actors~\cite{ng2022coordinated,saeed2022trollmagnifier}, it is not applicable to content spread autonomously and in good faith by regular users, for example in the wake of the uncertainty surrounding the COVID-19 pandemic.
In the case of crisis scenarios, misleading content with significant risk factors might proliferate as many different variants of the original content as a consequence of the broad and diverse vector of sharing by people.
This might render the initial set of seed hashes and similarity threshold being used ineffective in tracing the spread of the images.
One such example of this was the spreading of videos of the Christchurch mosque shooting, where Facebook reported that their systems were defeated by variants of the original videos as of bad actors started sharing it~\cite{newsroom_2016}.
Platforms could use methods from online learning to update the seed set of hashes to flag, and dynamically adjust the distance threshold in specific crisis cases to better handle such events.

Apart from applying soft moderation warning labels, Twitter outlines ``reduced visibility'' as a possible consequence for tweets violating their Civic Integrity Policy~\cite{twitter_civic_info}.
This means it is possible Twitter could have identified all of the images subsequently identified by \systemname, and chosen to reduce visibility of tweets including them instead of applying warning labels.
However, we cannot analyze this phenomenon due to lack of any public metrics regarding visibility of tweets. 
Due to the nature of the evaluation datasets, images containing multi-lingual text could not be evaluated.
Therefore, future applications of the system in multi-lingual settings might require further configuration of the underlying OCR engine (e.g. specifying \textit{``languageHints''} parameters in Cloud Vision OCR engine with the intended language of evaluation).

\descr{Future work.}
In the future, we plan to apply \systemname to other online platforms.
The approach is platform-independent, and this could give interesting insights on how misinformation spreads across different online communities, and which communities are particularly influential in generating viral image misinformation.
For example, we believe quite strongly that because of its low computational overhead and high performance, it makes a good fit for deployment by decentralized social networks (e.g., Mastodon), an environment where recent work has shown the potential to actually help \emph{improve} \systemname's performance via federated model sharing~\cite{mastodontoxic}.
Since our approach does not require any platform-specific information, it is particularly interesting as a basis for further research, especially at a time where academic research is being seriously threatened by the discontinuation of the Twitter Academic API as we know it.
Regardless of the specifics of future work, our presentation of \systemname provides a roadmap for measuring, tuning, and benchmarking soft moderation systems; critical for any moderation tool's success.
We strongly believe that the computer security community has a lot to say in this space, and hope that more researchers will get into this space.

\descr{Acknowledgments.} We would like to thank the anonymous reviewers and the anonymous shepherd for their feedback, and Primah Muwanga for her help setting up the \systemname artifact after the paper was accepted.
This work was supported by the NSF under grants CNS-1942610, CNS-2114407, CNS-2114411, CNS-2247867, CNS-2247868, and IIS-2046590, and by a grant from the Media Ecosystems Analysis Group (MEAG).

\begin{small}
\bibliographystyle{abbrv}
\bibliography{refs.bib}

\begin{thebibliography}{10}

\bibitem{kiely_2020}
Trump's repeated false attacks on mail-in ballots.
\newblock
  \url{https://www.factcheck.org/2020/09/trumps-repeated-false-attacks-on-mail-in-ballots/},
  Sep 2020.

\bibitem{abdali2021identifying}
S.~Abdali, R.~Gurav, S.~Menon, D.~Fonseca, N.~Entezari, N.~Shah, and E.~E.
  Papalexakis.
\newblock Identifying misinformation from website screenshots.
\newblock {\em arXiv preprint arXiv:2102.07849}, 2021.

\bibitem{abdelnabi2022open}
S.~Abdelnabi, R.~Hasan, and M.~Fritz.
\newblock Open-domain, content-based, multi-modal fact-checking of
  out-of-context images via online resources.
\newblock In {\em Proceedings of the IEEE/CVF Conference on Computer Vision and
  Pattern Recognition}, pages 14940--14949, 2022.

\bibitem{abilov2021voterfraud2020}
A.~Abilov, Y.~Hua, H.~Matatov, O.~Amir, and M.~Naaman.
\newblock Voterfraud2020: a multi-modal dataset of election fraud claims on
  twitter.
\newblock In {\em Proceedings of the International AAAI Conference on Web and
  Social Media}, volume~15, 2021.

\bibitem{abril_2020}
D.~Abril.
\newblock Facebook reveals that massive amounts of misinformation flooded its
  service during the election, Nov 2020.

\bibitem{ahmed2010secure}
F.~Ahmed, M.~Y. Siyal, and V.~U. Abbas.
\newblock A secure and robust hash-based scheme for image authentication.
\newblock {\em Signal Processing}, 90(5), 2010.

\bibitem{ahmed2019content}
K.~T. Ahmed, S.~Ummesafi, and A.~Iqbal.
\newblock Content based image retrieval using image features information
  fusion.
\newblock {\em Information Fusion}, 51, 2019.

\bibitem{alkhowaiter2022evaluating}
M.~Alkhowaiter, K.~Almubarak, and C.~Zou.
\newblock Evaluating perceptual hashing algorithms in detecting image
  manipulation over social media platforms.
\newblock In {\em 2022 IEEE International Conference on Cyber Security and
  Resilience (CSR)}. IEEE, 2022.

\bibitem{alvarado_2020}
E.~Alvarado, D.~A. Graham, C.~Murphy, and A.~Amy Weiss-Meyer.
\newblock The bush-gore recount is an omen for 2020.
\newblock
  \url{https://www.theatlantic.com/politics/archive/2020/08/bush-gore-florida-recount-oral-history/614404/},
  Aug 2020.

\bibitem{aneja2021cosmos}
S.~Aneja, C.~Bregler, and M.~Nießner.
\newblock Cosmos: Catching out-of-context misinformation with self-supervised
  learning, 2021.

\bibitem{arsht2018human}
A.~Arsht and D.~Etcovitch.
\newblock The human cost of online content moderation.
\newblock {\em Harvard Journal of Law and Technology}, 2018.

\bibitem{bayar2016deep}
B.~Bayar and M.~C. Stamm.
\newblock A deep learning approach to universal image manipulation detection
  using a new convolutional layer.
\newblock In {\em {ACM IH}}, 2016.

\bibitem{mastodontoxic}
H.~Bin~Zia, A.~Raman, I.~Castro, I.~Hassan~Anaobi, E.~De~Cristofaro, N.~Sastry,
  and G.~Tyson.
\newblock Toxicity in the decentralized web and the potential for model
  sharing.
\newblock {\em Proceedings of the ACM on Measurement and Analysis of Computing
  Systems}, 6(2), jun 2022.

\bibitem{bond_2020}
S.~Bond.
\newblock Twitter expands warning labels to slow spread of election
  misinformation, Oct 2020.

\bibitem{lawfare_2022}
S.~Bradshaw and S.~Grossman.
\newblock {Were Facebook and Twitter consistent in labeling misleading posts
  during the 2020 election?}
\newblock
  \url{https://www.lawfareblog.com/were-facebook-and-twitter-consistent-labeling-misleading-posts-during-2020-election},
  2022.

\bibitem{usatoday_410}
C.~Caldera.
\newblock Fact check: Map showing trump landslide based on false report of
  seized election servers in germany.
\newblock
  \url{https://www.usatoday.com/story/news/factcheck/2020/11/18/fact-check-fake-map-shows-trump-with-410-electoral-votes/3767048001/},
  Nov 2020.

\bibitem{calma_2021}
J.~Calma.
\newblock Facebook will add a new label to some climate change posts in the uk,
  Feb 2021.

\bibitem{chen2020uniter}
Y.-C. Chen, L.~Li, L.~Yu, A.~El~Kholy, F.~Ahmed, Z.~Gan, Y.~Cheng, and J.~Liu.
\newblock Uniter: Universal image-text representation learning.
\newblock In {\em European conference on computer vision}. Springer, 2020.

\bibitem{conger_2020}
K.~Conger.
\newblock Twitter says it labeled 0.2
  disputed., Nov 2020.

\bibitem{connellan_2021}
S.~Connellan.
\newblock Facebook to add labels to climate change posts, Oct 2021.

\bibitem{culliford_2020}
E.~Culliford.
\newblock Twitter launches labels, warnings on misleading covid-19 information.
\newblock
  \url{https://www.reuters.com/article/us-health-coronavirus-twitter/twitter-launches-labels-warnings-on-misleading-covid-19-information-idUSKBN22N2E4},
  May 2020.

\bibitem{zoom_csam}
E.~Culliford.
\newblock Exclusive zoom has joined tech industry counterterrorism group, Dec
  2021.

\bibitem{culliford_2021}
E.~Culliford.
\newblock Facebook to label all posts about covid-19 vaccines.
\newblock
  \url{https://www.reuters.com/article/us-health-coronavirus-facebook/facebook-to-label-all-posts-about-covid-19-vaccines-idUSKBN2B70NJ},
  Mar 2021.

\bibitem{davis2019open}
A.~Davis and G.~Rosen.
\newblock Open-sourcing photo-and video-matching technology to make the
  internet safer.
\newblock {\em Facebook Newsroom}, 2019.

\bibitem{dewan2017towards}
P.~Dewan, A.~Suri, V.~Bharadhwaj, A.~Mithal, and P.~Kumaraguru.
\newblock {Towards Understanding Crisis Events On Online Social Networks
  Through Pictures}.
\newblock In {\em {ASONAM}}, 2017.

\bibitem{dutta2016vgg}
A.~Dutta, A.~Gupta, and A.~Zissermann.
\newblock Vgg image annotator (via).
\newblock {\em URL: http://www. robots. ox. ac. uk/vgg/software/via}, 2, 2016.

\bibitem{ester1996density}
M.~Ester, H.-P. Kriegel, J.~Sander, X.~Xu, et~al.
\newblock A density-based algorithm for discovering clusters in large spatial
  databases with noise.
\newblock In {\em kdd}, volume~96, 1996.

\bibitem{garimella2017image}
K.~Garimella and D.~Eckles.
\newblock Image based misinformation on whatsapp.
\newblock In {\em {International AAAI Conference on Web and Social Media
  (ICWSM)}}, 2017.

\bibitem{garimella2020images}
K.~Garimella and D.~Eckles.
\newblock Images and misinformation in political groups: Evidence from whatsapp
  in india.
\newblock {\em arXiv:2005.09784}, 2020.

\bibitem{goga2015doppelganger}
O.~Goga, G.~Venkatadri, and K.~P. Gummadi.
\newblock The doppelg{\"a}nger bot attack: Exploring identity impersonation in
  online social networks.
\newblock In {\em Proceedings of the 2015 internet measurement conference},
  2015.

\bibitem{gonzalez2023understanding}
F.~Gonz{\'a}lez-Pizarro and S.~Zannettou.
\newblock Understanding and detecting hateful content using contrastive
  learning.
\newblock In {\em Proceedings of the International AAAI Conference on Web and
  Social Media}, volume~17, 2023.

\bibitem{google_cloud_vision}
Google.
\newblock Cloud vision api.
\newblock \url{https://cloud.google.com/vision/}.

\bibitem{google_csam}
Google.
\newblock Google’s efforts to combat online child sexual abuse material.
\newblock
  \url{https://transparencyreport.google.com/child-sexual-abuse-material/reporting}.

\bibitem{graham_rodriguez_2020}
M.~Graham and S.~Rodriguez.
\newblock Twitter and facebook race to label a slew of posts making false
  election claims before all votes counted, Nov 2020.

\bibitem{gudivada1995content}
V.~N. Gudivada and V.~V. Raghavan.
\newblock Content based image retrieval systems.
\newblock {\em Computer}, 28(9), 1995.

\bibitem{hao2021s}
Q.~Hao, L.~Luo, S.~T. Jan, and G.~Wang.
\newblock It's not what it looks like: Manipulating perceptual hashing based
  applications.
\newblock In {\em Proceedings of the 2021 ACM SIGSAC Conference on Computer and
  Communications Security}, 2021.

\bibitem{he2016deep}
K.~He, X.~Zhang, S.~Ren, and J.~Sun.
\newblock Deep residual learning for image recognition.
\newblock In {\em Proceedings of the IEEE conference on computer vision and
  pattern recognition}, 2016.

\bibitem{hu2022badhash}
S.~Hu, Z.~Zhou, Y.~Zhang, L.~Y. Zhang, Y.~Zheng, Y.~He, and H.~Jin.
\newblock Badhash: Invisible backdoor attacks against deep hashing with clean
  label.
\newblock In {\em Proceedings of the 30th ACM International Conference on
  Multimedia}, 2022.

\bibitem{hunt2020misinformation}
K.~Hunt, B.~Wang, and J.~Zhuang.
\newblock Misinformation debunking and cross-platform information sharing
  through twitter during hurricanes harvey and irma: a case study on shelters
  and id checks.
\newblock {\em Natural Hazards}, 103(1), 2020.

\bibitem{ith2015microsoft}
T.~Ith.
\newblock Microsoft’s photodna: Protecting children and businesses in the
  cloud.
\newblock {\em URL: https://news. microsoft.
  com/features/microsofts-photodna-protecting-children-andbusinesses-in-the-cloud/366
  REFERENCES}, 2015.

\bibitem{javed2020first}
R.~T. Javed, M.~E. Shuja, M.~Usama, J.~Qadir, W.~Iqbal, G.~Tyson, I.~Castro,
  and K.~Garimella.
\newblock A first look at covid-19 messages on whatsapp in pakistan.
\newblock In {\em 2020 IEEE/ACM International Conference on Advances in Social
  Networks Analysis and Mining (ASONAM)}. IEEE, 2020.

\bibitem{jin2016novel}
Z.~Jin, J.~Cao, Y.~Zhang, J.~Zhou, and Q.~Tian.
\newblock Novel visual and statistical image features for microblogs news
  verification.
\newblock {\em IEEE transactions on multimedia}, 19(3):598--608, 2016.

\bibitem{johnson2019billion}
J.~Johnson, M.~Douze, and H.~J{\'e}gou.
\newblock Billion-scale similarity search with {GPUs}.
\newblock {\em IEEE Transactions on Big Data}, 7(3), 2019.

\bibitem{kazemi2021tiplines}
A.~Kazemi, K.~Garimella, G.~K. Shahi, D.~Gaffney, and S.~A. Hale.
\newblock Tiplines to combat misinformation on encrypted platforms: a case
  study of the 2019 indian election on whatsapp.
\newblock {\em arXiv preprint arXiv:2106.04726}, 2021.

\bibitem{kenneally2012menlo}
E.~Kenneally and D.~Dittrich.
\newblock The menlo report: Ethical principles guiding information and
  communication technology research.
\newblock {\em Available at SSRN 2445102}, 2012.

\bibitem{kharraz2018surveylance}
A.~Kharraz, W.~Robertson, and E.~Kirda.
\newblock Surveylance: Automatically detecting online survey scams.
\newblock In {\em 2018 IEEE Symposium on Security and Privacy (SP)}. IEEE,
  2018.

\bibitem{kintis2017hiding}
P.~Kintis, N.~Miramirkhani, C.~Lever, Y.~Chen, R.~Romero-G{\'o}mez,
  N.~Pitropakis, N.~Nikiforakis, and M.~Antonakakis.
\newblock Hiding in plain sight: A longitudinal study of combosquatting abuse.
\newblock In {\em Proceedings of the 2017 ACM SIGSAC Conference on Computer and
  Communications Security}, 2017.

\bibitem{lange_2020}
J.~Lange.
\newblock {Twitter is now flagging the use of 'oxygen' and 'frequency' in the
  same tweet, prompting new meme}.
\newblock
  \url{https://theweek.com/speedreads/922275/twitter-now-flagging-use-oxygen-frequency-same-tweet-prompting-new-meme},
  2020.

\bibitem{li2019visualbert}
L.~H. Li, M.~Yatskar, D.~Yin, C.-J. Hsieh, and K.-W. Chang.
\newblock Visualbert: A simple and performant baseline for vision and language.
\newblock {\em arXiv preprint arXiv:1908.03557}, 2019.

\bibitem{lin2014microsoft}
T.-Y. Lin, M.~Maire, S.~Belongie, J.~Hays, P.~Perona, D.~Ramanan,
  P.~Doll{\'a}r, and C.~L. Zitnick.
\newblock Microsoft coco: Common objects in context.
\newblock In {\em Computer Vision--ECCV 2014: 13th European Conference, Zurich,
  Switzerland, September 6-12, 2014, Proceedings, Part V 13}. Springer, 2014.

\bibitem{lowe1999object}
D.~G. Lowe.
\newblock Object recognition from local scale-invariant features.
\newblock In {\em Proceedings of the seventh IEEE international conference on
  computer vision}, volume~2. Ieee, 1999.

\bibitem{lyons_2020}
K.~Lyons.
\newblock {Twitter promises to fine-tune its 5G coronavirus labeling after
  unrelated tweets were flagged}.
\newblock
  \url{https://www.theverge.com/2020/6/27/21305503/twitter-labels-5g-conspiracy-coronavirus},
  2020.

\bibitem{matatov2018dejavu}
H.~Matatov, A.~Bechhofer, L.~Aroyo, O.~Amir, and M.~Naaman.
\newblock Dejavu: a system for journalists to collaboratively address visual
  misinformation.
\newblock In {\em Computation+ Journalism Symposium. Miami}, 2018.

\bibitem{matatov2022stop}
H.~Matatov, M.~Naaman, and O.~Amir.
\newblock Stop the [image] steal: The role and dynamics of visual content in
  the 2020 us election misinformation campaign.
\newblock {\em arXiv preprint arXiv:2209.02007}, 2022.

\bibitem{mchugh2012interrater}
M.~L. McHugh.
\newblock Interrater reliability: the kappa statistic.
\newblock {\em Biochemia medica}, 2012.

\bibitem{miramirkhani2016dial}
N.~Miramirkhani, O.~Starov, and N.~Nikiforakis.
\newblock Dial one for scam: A large-scale analysis of technical support scams.
\newblock {\em arXiv preprint arXiv:1607.06891}, 2016.

\bibitem{monga2006perceptual}
V.~Monga and B.~L. Evans.
\newblock {Perceptual Image Hashing Via Feature Points: Performance Evaluation
  and Tradeoffs}.
\newblock {\em IEEE Transactions on Image Processing}, 2006.

\bibitem{newsroom_2016}
F.~Newsroom.
\newblock Partnering to help curb spread of online terrorist content, Dec 2016.

\bibitem{ng2022coordinated}
L.~H.~X. Ng, J.~Moffitt, and K.~M. Carley.
\newblock Coordinated through aweb of images: Analysis of image-based influence
  operations from china, iran, russia, and venezuela.
\newblock {\em arXiv preprint arXiv:2206.03576}, 2022.

\bibitem{nikiforakis2014stranger}
N.~Nikiforakis, F.~Maggi, G.~Stringhini, M.~Z. Rafique, W.~Joosen, C.~Kruegel,
  F.~Piessens, G.~Vigna, and S.~Zanero.
\newblock Stranger danger: exploring the ecosystem of ad-based url shortening
  services.
\newblock In {\em Proceedings of the 23rd international conference on World
  wide web}, 2014.

\bibitem{paudel2022lambretta}
P.~Paudel, J.~Blackburn, E.~De~Cristofaro, S.~Zannettou, and G.~Stringhini.
\newblock Lambretta: Learning to rank for twitter soft moderation.
\newblock In {\em {IEEE Symposium on Security and Privacy}}, 2023.

\bibitem{tweet_annotations}
T.~D. Platform.
\newblock Tweet annotations.
\newblock
  \url{https://developer.twitter.com/en/docs/twitter-api/annotations/overview}.

\bibitem{qu2023evolution}
Y.~Qu, X.~He, S.~Pierson, M.~Backes, Y.~Zhang, and S.~Zannettou.
\newblock On the evolution of (hateful) memes by means of multimodal
  contrastive learning.
\newblock In {\em 2023 IEEE Symposium on Security and Privacy (SP)}. IEEE,
  2023.

\bibitem{radford2021learning}
A.~Radford, J.~W. Kim, C.~Hallacy, A.~Ramesh, G.~Goh, S.~Agarwal, G.~Sastry,
  A.~Askell, P.~Mishkin, J.~Clark, et~al.
\newblock Learning transferable visual models from natural language
  supervision.
\newblock In {\em International conference on machine learning}. PMLR, 2021.

\bibitem{reimers2019sentence}
N.~Reimers and I.~Gurevych.
\newblock Sentence-bert: Sentence embeddings using siamese bert-networks.
\newblock In {\em Proceedings of the 2019 Conference on Empirical Methods in
  Natural Language Processing and the 9th International Joint Conference on
  Natural Language Processing (EMNLP-IJCNLP)}, 2019.

\bibitem{reis2020can}
J.~Reis, P.~d.~F. Melo, K.~Garimella, and F.~Benevenuto.
\newblock Can whatsapp benefit from debunked fact-checked stories to reduce
  misinformation?
\newblock {\em arXiv preprint arXiv:2006.02471}, 2020.

\bibitem{romo_2020}
V.~Romo.
\newblock Twitter to remove or place warning labels on covid vaccine conspiracy
  tweets.
\newblock
  \url{https://www.npr.org/sections/coronavirus-live-updates/2020/12/16/947355414/twitter-to-remove-or-place-warning-labels-on-covid-vaccine-conspiracy-tweets},
  Dec 2020.

\bibitem{rosen_2021}
G.~Rosen.
\newblock An update on our work to keep people informed and limit
  misinformation about covid-19.
\newblock \url{https://about.fb.com/news/2020/04/covid-19-misinfo-update/}, May
  2021.

\bibitem{rublee2011orb}
E.~Rublee, V.~Rabaud, K.~Konolige, and G.~Bradski.
\newblock Orb: An efficient alternative to sift or surf.
\newblock In {\em 2011 International conference on computer vision}. Ieee,
  2011.

\bibitem{saeed2022trollmagnifier}
M.~H. Saeed, S.~Ali, J.~Blackburn, E.~De~Cristofaro, S.~Zannettou, and
  G.~Stringhini.
\newblock Trollmagnifier: Detecting state-sponsored troll accounts on reddit.
\newblock In {\em IEEE Symposium on Security and Privacy (SP)}, 2022.

\bibitem{twitternumbers}
D.~Sayce.
\newblock The number of tweets per day in 2022.
\newblock \url{https://www.dsayce.com/social-media/tweets-day/}.

\bibitem{simonyan2014very}
K.~Simonyan and A.~Zisserman.
\newblock Very deep convolutional networks for large-scale image recognition.
\newblock {\em arXiv preprint arXiv:1409.1556}, 2014.

\bibitem{sklan2015toward}
J.~E. Sklan, A.~J. Plassard, D.~Fabbri, and B.~A. Landman.
\newblock Toward content-based image retrieval with deep convolutional neural
  networks.
\newblock In {\em Medical Imaging 2015: Biomedical Applications in Molecular,
  Structural, and Functional Imaging}, volume 9417. SPIE, 2015.

\bibitem{staff_2020}
R.~Staff.
\newblock Fact check: Tabulation machines in arizona can read ballots marked
  with sharpie pens.
\newblock
  \url{https://www.reuters.com/article/uk-factcheck-sharpie-arizona/fact-check-tabulation-machines-in-arizona-can-read-ballots-marked-with-sharpie-pens-idUSKBN27L2R5},
  Nov 2020.

\bibitem{starbird2019disinformation}
K.~Starbird, A.~Arif, and T.~Wilson.
\newblock Disinformation as collaborative work: Surfacing the participatory
  nature of strategic information operations.
\newblock {\em Proceedings of the ACM on Human-Computer Interaction}, 3(CSCW),
  2019.

\bibitem{verge_platforms}
N.~Statt.
\newblock Major tech platforms say they’re ‘jointly combating fraud and
  misinformation’ about covid-19, Mar 2020.

\bibitem{steiger2021psychological}
M.~Steiger, T.~J. Bharucha, S.~Venkatagiri, M.~J. Riedl, and M.~Lease.
\newblock The psychological well-being of content moderators: the emotional
  labor of commercial moderation and avenues for improving support.
\newblock In {\em Proceedings of the 2021 CHI conference on human factors in
  computing systems}, 2021.

\bibitem{szegedy2016rethinking}
C.~Szegedy, V.~Vanhoucke, S.~Ioffe, J.~Shlens, and Z.~Wojna.
\newblock Rethinking the inception architecture for computer vision.
\newblock In {\em Proceedings of the IEEE conference on computer vision and
  pattern recognition}, 2016.

\bibitem{tang2008robust}
Z.~Tang, S.~Wang, X.~Zhang, W.~Wei, and S.~Su.
\newblock Robust image hashing for tamper detection using non-negative matrix
  factorization.
\newblock {\em Journal of ubiquitous convergence technology}, 2(1), 2008.

\bibitem{tineye}
TinEye.
\newblock Tineye: Reverse image search.
\newblock \url{https://tineye.com/}.

\bibitem{tola2009daisy}
E.~Tola, V.~Lepetit, and P.~Fua.
\newblock Daisy: An efficient dense descriptor applied to wide-baseline stereo.
\newblock {\em IEEE transactions on pattern analysis and machine intelligence},
  32(5), 2009.

\bibitem{twitter_manipulated_media}
Twitter.
\newblock {Synthetic and manipulated media policy}.
\newblock
  \url{https://help.twitter.com/en/rules-and-policies/manipulated-media}, 2020.

\bibitem{twitter_civic_info}
Twitter.
\newblock Twitter's civic integrity policy | twitter help, 2020.

\bibitem{wang2021milvus}
J.~Wang, X.~Yi, R.~Guo, H.~Jin, P.~Xu, S.~Li, X.~Wang, X.~Guo, C.~Li, X.~Xu,
  K.~Yu, Y.~Yuan, Y.~Zou, J.~Long, Y.~Cai, Z.~Li, Z.~Zhang, Y.~Mo, J.~Gu,
  R.~Jiang, Y.~Wei, and C.~Xie.
\newblock Milvus: A purpose-built vector data management system.
\newblock In {\em Proceedings of the 2021 International Conference on
  Management of Data}, 2021.

\bibitem{wang2015visual}
X.~Wang, K.~Pang, X.~Zhou, Y.~Zhou, L.~Li, and J.~Xue.
\newblock A visual model-based perceptual image hash for content
  authentication.
\newblock {\em IEEE Transactions on Information Forensics and Security}, 10(7),
  2015.

\bibitem{wang2021targeted}
X.~Wang, Z.~Zhang, G.~Lu, and Y.~Xu.
\newblock Targeted attack and defense for deep hashing.
\newblock In {\em Proceedings of the 44th International ACM SIGIR Conference on
  Research and Development in Information Retrieval}, 2021.

\bibitem{wang2023understanding}
Y.~Wang, C.~Ling, and G.~Stringhini.
\newblock Understanding the use of images to spread covid-19 misinformation on
  twitter.
\newblock {\em {Proceedings of the ACM in Human Computer Interaction (CSCW)}},
  2023.

\bibitem{wang2020understanding}
Y.~Wang, F.~Tamahsbi, J.~Blackburn, B.~Bradlyn, E.~De~Cristofaro, D.~Magerman,
  S.~Zannettou, and G.~Stringhini.
\newblock Understanding the use of fauxtography on social media.
\newblock In {\em ICWSM}, 2021.

\bibitem{wilson2020cross}
T.~Wilson and K.~Starbird.
\newblock Cross-platform disinformation campaigns: lessons learned and next
  steps.
\newblock {\em Harvard Kennedy School Misinformation Review}, 1(1), 2020.

\bibitem{zannettou2021won}
S.~Zannettou.
\newblock I won the election: An empirical analysis of soft moderation
  interventions on twitter.
\newblock In {\em Proceedings of the International AAAI Conference on Web and
  Social Media}, volume~15, 2021.

\bibitem{zannettou2018origins}
S.~Zannettou, T.~Caulfield, J.~Blackburn, E.~De~Cristofaro, M.~Sirivianos,
  G.~Stringhini, and G.~Suarez-Tangil.
\newblock {On the Origins of Memes by Means of Fringe Web Communities}.
\newblock In {\em Proceedings of the Internet Measurement Conference 2018},
  2018.

\bibitem{zannettou2019characterizing}
S.~Zannettou, T.~Caulfield, B.~Bradlyn, E.~De~Cristofaro, G.~Stringhini, and
  J.~Blackburn.
\newblock Characterizing the use of images in state-sponsored information
  warfare operations by russian trolls on twitter.
\newblock In {\em Proceedings of the international AAAI conference on web and
  social media}, 2020.

\bibitem{zhao2010perceptual}
Y.~Zhao and W.~Wei.
\newblock Perceptual image hash for tampering detection using zernike moments.
\newblock In {\em 2010 IEEE International Conference on Progress in Informatics
  and Computing}, volume~2. IEEE, 2010.

\bibitem{zhou2018learning}
P.~Zhou, X.~Han, V.~I. Morariu, and L.~S. Davis.
\newblock Learning rich features for image manipulation detection.
\newblock In {\em Proceedings of the IEEE Conference on Computer Vision and
  Pattern Recognition}, pages 1053--1061, 2018.

\bibitem{zlatkova2019fact}
D.~Zlatkova, P.~Nakov, and I.~Koychev.
\newblock Fact-checking meets fauxtography: Verifying claims about images.
\newblock In {\em EMNLP-IJCNLP}, 2019.

\end{thebibliography}
\end{small}

\begin{appendix}

\section{Twitter's content policies and violations.}
\label{sec:appendix_policy}
We list the four different categories of Twitter's platform policy violations and the corresponding rules for each category below.

\begin{itemize}[topsep=0pt,parsep=0pt,partopsep=0ptnoitemsep,topsep=0pt,parsep=0pt,partopsep=0pt]
    
    \item Misleading information about how to participate.

        \begin{itemize}

            \item Images misleading people about participation procedures and requirements.

            \item Images sowing confusion about officials and institutions.

            \item Images discussing threats on voting locations.

        \end{itemize}

    \item Misleading information intended to intimidate people from civic processes.

     \begin{itemize}

            \item Images about votes not being counted.

            \item Images about Equipment Problems.

            \item Images about disruptions at voting locations.

            \item Images about closing of polls.

        \end{itemize}

        \item Misleading information about outcomes of civic processes.

    \begin{itemize}

            \item Images undermining public confidence in methods and results of election.

            \item Images with misleading claims about election rigging.

            \item Images with misleading claims about ballot tampering.

            \item Images with misleading claims about vote tallying.

            \item Images with declaration of premature victory.

            \item Images casting doubt on outcome of civic processes.

            \item Images calling for interference with the implementation of election results.

        \end{itemize}

    \item Synthetic and Manipulated Media.

    \begin{itemize}

            \item Images that are significantly and deceptively altered, manipulated or fabricated.

            \item Images shared with malicious intent , including out of context tweets sharing the media.

        \end{itemize}

\end{itemize}

\end{appendix}





\end{document}